%% file: acl_latex.tex
\documentclass[11pt]{article}

\usepackage[preprint]{acl}

\usepackage{times}
\usepackage{latexsym}

\usepackage[T1]{fontenc}

\usepackage[utf8]{inputenc}

\usepackage{microtype}

\usepackage{inconsolata}

\usepackage{graphicx}

\usepackage{booktabs}
\usepackage{amsfonts}
\usepackage{tcolorbox}
\tcbuselibrary{skins,breakable,listings}

\usepackage{listings}
\usepackage[table]{xcolor}
\usepackage{amsmath}
\usepackage{pifont}
\usetikzlibrary{arrows.meta,calc,fit,positioning}

\newcommand{\sparagraph}[1]{\vspace{3pt}\par\noindent\textbf{#1}}

\title{Learning When to Translate for Multilingual Reasoning}

\author{
  Deokhyung Kang\textsuperscript{1}, Hyounghun Kim\textsuperscript{1,2}\\
  \textsuperscript{1}Graduate School of Artificial Intelligence\\
  \textsuperscript{2}Department of Computer Science \& Engineering\\
  POSTECH, South Korea \\
  \texttt{\{deokhk, h.kim\}@postech.ac.kr} \\}

\begin{document}
\maketitle
\input{latex/content/0_abstract}

\input{latex/content/1_introduction}
\input{latex/content/2_related_works}
\input{latex/content/3_method}

\input{latex/content/4_experiments}
\input{latex/content/5_conclusion}
\input{latex/content/limitations}
\input{latex/content/ethical_considerations}
\input{latex/content/acknowledgement}

\bibliography{custom}

\appendix
\clearpage
\input{latex/appendix/appendix_section_A}
\input{latex/appendix/appendix_section_B}
\input{latex/appendix/appendix_section_C}
\input{latex/appendix/appendix_section_D}

\input{latex/appendix/appendix_section_E}
\input{latex/appendix/appendix_section_F}

\end{document}

%% file: latex/content/0_abstract.tex
\begin{abstract}
Reasoning language models (RLMs) achieve strong performance on complex reasoning tasks, but still exhibit substantial multilingual reasoning gaps, largely due to language-understanding failures when interpreting non-English inputs. Translation into English can mitigate these failures by expressing non-English inputs in a form that RLMs can more reliably interpret, yet translating every input incurs unnecessary inference cost when the model can reason reliably from the original query. To address this challenge, we propose \textsc{Luar}, a \textbf{L}anguage \textbf{U}nderstanding Boundary-\textbf{a}ware \textbf{R}einforcement Learning framework that trains RLMs to selectively invoke translation when their direct understanding is unreliable. \textsc{Luar} trains the model to either solve the original input directly or reason over its English translation, encouraging translation only when translator-augmented reasoning is expected to substantially outperform direct reasoning. Across multilingual reasoning benchmarks, \textsc{Luar} outperforms training-based baselines, with particularly large gains on low-resource languages. On Qwen3-4B, it improves average accuracy by 8.4 percentage points over standard GRPO. Further analysis shows that \textsc{Luar} concentrates translator calls on queries where translation benefits reasoning, helping to reduce inference cost.\footnote{The project will be made publicly available at \url{https://github.com/deokhk/LUAR}.}

\end{abstract}

%% file: latex/content/1_introduction.tex
\begin{figure}[!t]
\centering
\includegraphics[width=\columnwidth]{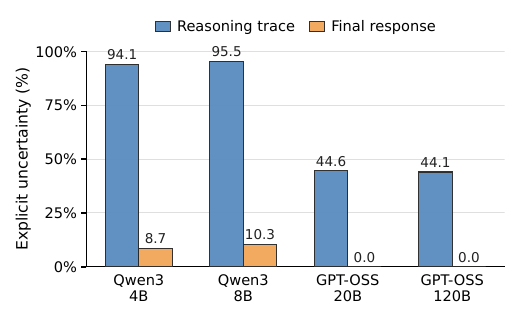}
\caption{Explicit uncertainty rates of RLMs on PolyMath-Low. We report the percentage of query-understanding failures where the model explicitly expresses uncertainty about the meaning of the query, measured separately in reasoning traces and final responses. Evaluation details are provided in Appendix~\ref{app:verbalized-confusion-evaluation}.}
\label{fig:verbalized_failure_rates}

\end{figure}

\section{Introduction}

Reasoning language models (\textbf{RLMs}) have made rapid progress on complex reasoning tasks by generating intermediate reasoning traces before producing final responses~\cite{jaech2024openai, deepseekai2025deepseekr1incentivizingreasoningcapability, yang2025qwen3technicalreport}. Despite this progress, RLMs still exhibit multilingual reasoning performance gaps~\cite{huang-etal-2025-benchmax, wang2025polymath}. In particular, they perform much better on queries in high-resource languages such as English than on those in low-resource languages, making RLMs less reliable for speakers of underrepresented languages.

Recent studies~\cite{kang2026multilingualreasoninggapsemerge, huang2026tapo} suggest that these gaps are largely driven by failures in language understanding, especially for low-resource languages. The key bottleneck is therefore often not reasoning itself, but accurately comprehending the meaning of the input. A straightforward approach is to directly improve the model's language understanding across languages. However, this approach is difficult to scale across languages, especially for RLMs. Continual training on target-language data~\cite{zhu-etal-2024-question, ustun2024aya, ko-etal-2025-understand} requires targeted training data and optimization for each language. Model merging with multilingual models reduces language-specific data requirements~\cite{yoon-etal-2024-langbridge, huang2024mindmerger, pipatanakul2025adapting}, but requires careful adaptation to post-trained RLMs to avoid reasoning performance degradation~\cite{yang2025rcp}. These limitations motivate our central research question: rather than directly improving the model's own language understanding across languages, \textbf{can RLMs learn to recognize when an input falls outside their own language-understanding boundary and selectively invoke translation?}

To examine whether current RLMs exhibit signals relevant to this boundary, we analyze cases where incorrect answers stem from failures to understand the query. We measure the explicit uncertainty rate, the percentage of such cases where the model explicitly expresses uncertainty about its understanding. Figure~\ref{fig:verbalized_failure_rates} shows that models often reveal such uncertainty in reasoning traces, yet still produce confident final responses. This suggests that RLMs exhibit internal signals of language-understanding failure, but lack a mechanism to act on them. This motivates the idea of invoking English translation when needed, which can present the input in a language the model understands better.

To operationalize this idea, we propose \textsc{Luar}, a \textbf{L}anguage \textbf{U}nderstanding Boundary-\textbf{a}ware \textbf{R}einforcement Learning framework that teaches RLMs to invoke translation only when their direct understanding is unreliable. \textsc{Luar} trains this decision inside the reasoning trace through two stages. It first uses translator-call supervised fine-tuning to expose the model to both direct and translator-augmented reasoning. This is followed by boundary-aware online reinforcement learning with GRPO~\cite{shao2024deepseekmathpushinglimitsmathematical}, which rewards translator calls only when translation is expected to substantially improve final-answer accuracy.

Across multilingual mathematical and STEM reasoning benchmarks~\cite{wang2025polymath,xuanmmluprox}, \textsc{Luar} improves Qwen3-4B and Qwen3-8B~\cite{yang2025qwen3technicalreport} over standard GRPO and other training-based baselines, with especially large gains on low-resource languages. Its learned policy calls the translator more often for queries that benefit from translation, including those in unseen low-resource languages such as Yoruba and Zulu.

We summarize our main contributions as follows: 
(i) We introduce a view of multilingual reasoning centered on language-understanding boundaries, in which RLMs selectively invoke translation when direct understanding is unreliable. (ii) We propose \textsc{Luar}, a two-stage framework that learns when to invoke translation from outcome-based boundary signals. (iii) We demonstrate that \textsc{Luar} substantially improves multilingual reasoning, especially for low-resource languages, and reduces inference cost.

%% file: latex/content/2_related_works.tex
\section{Related Work}
Despite the rapid progress of language models on complex reasoning tasks, they continue to exhibit multilingual reasoning gaps, especially for low-resource languages~\cite{huang-etal-2025-benchmax,barua2026long,ki2026makesgoodmultilingualreasoning}.
Prior work has aimed to reduce these gaps through test-time translation~\cite{shilanguage,liu-etal-2025-translation}, model merging~\cite{yoon-etal-2024-langbridge,huang2024mindmerger}, and training-based methods~\cite{zhu-etal-2024-question,she-etal-2024-mapo,chai2024xcotcrosslingualinstructiontuning,ko-etal-2025-understand,hwang2025learn,liu2026crosslingual}. With the emergence of RLMs, recent work has further examined how multilingual inputs are processed within their reasoning traces. These analyses show that RLMs often produce English-centric reasoning traces even for non-English inputs, and that reasoning in English is associated with stronger performance~\cite{yong2025crosslingual-test-time-scaling,Tam2025LanguageMH,qi-etal-2025-models}. \citet{kang2026multilingualreasoninggapsemerge} further show that understanding failures---i.e., failures to implicitly comprehend or translate the input into the trace language---are a major source of multilingual reasoning gaps in RLMs. To mitigate these failures, they propose Selective Translation, which uses an external detector to invoke translation when such understanding failures are predicted. Other recent methods aim to improve multilingual understanding jointly with reasoning ability during optimization~\cite{liu2026self,huang2026tapo}. 

While effective, these approaches either rely on a separately trained detector, which can introduce additional latency, or remain bounded by the model's own multilingual understanding, limiting gains on low-resource inputs. \textsc{Luar} addresses this gap by training the RLM itself to invoke an external translator when direct reasoning is unreliable, making translation an explicit action during reasoning.

%% file: latex/content/3_method.tex
\section{Method}
\label{sec:method}
\textsc{Luar} trains an RLM to selectively invoke translation when the input is likely to fall outside its language-understanding boundary. We first introduce an outcome-based surrogate for this boundary (\S\ref{sec:boundary-estimation}), and then describe \textsc{Luar}'s two-stage training procedure (\S\ref{sec:luar}). Figure~\ref{fig:luar-overview} provides an overview of \textsc{Luar}.
\subsection{Outcome-Based Boundary Estimation}
\label{sec:boundary-estimation}
\begin{figure}[!t]
\centering
\begin{tcolorbox}[
    colback=gray!5,
    colframe=gray!40,
    boxrule=0.5pt,
    arc=2pt,
    left=4pt,
    right=4pt,
    top=4pt,
    bottom=4pt,
    width=\columnwidth
]
\small
\texttt{<think>}\\
Okay, let's see. As I read the question, some parts are difficult for me to fully understand. I'll translate it into English to avoid any misunderstanding. Let me call the translator to get an English version of the question:\\
\texttt{<translator\_call>}\\
\{\textit{question in the original language}\}\\
\texttt{</translator\_call>}\\
\texttt{<translator\_response>}\\
\{\textit{ground-truth English reference translation}\}\\
\texttt{</translator\_response>}\\
Now that I have the translated version, I can proceed with the reasoning step by step.
\end{tcolorbox}

\caption{Translator-call prefix prepended to the reasoning trace for translator-augmented reasoning.}

\label{fig:translator-prefix}

\end{figure}

\input{latex/figure/main_figure}

Our goal is to identify cases where invoking translation is more reliable than directly reasoning over the original-language query. We therefore adopt an \textit{operational} definition of the model's language-understanding boundary: \textbf{a query falls outside this boundary when access to an English translation substantially improves the model's ability to solve it.} We estimate this by sampling responses from two reasoning modes on the same query. 
For each multilingual query $q$, \emph{direct reasoning} samples are generated by solving the original-language query without translation, while \emph{translator-augmented reasoning} samples are generated by prepending an English \textbf{reference} translation to the reasoning trace using the translator-call prefix in Figure~\ref{fig:translator-prefix}. This intervention conditions the reasoning trace on the intended English meaning of the input while avoiding translation noise.
We sample $N/2$ responses from each mode, with $N=8$ in our experiments.
Let $d_{\mathrm{acc}}(q)$ and $t_{\mathrm{acc}}(q)$ be the fractions of correct responses under direct and translator-augmented reasoning, respectively.
We define an outcome-based translation-usefulness label $u(q)\in\{0,1\}$ as
\begin{equation}
u(q) =
\mathbb{I}\left[
t_{\mathrm{acc}}(q) \ge d_{\mathrm{acc}}(q) + \delta
\right],
\label{eq:boundary-label}
\end{equation}
where $u(q)=1$ means that $q$ is estimated to fall outside the model's language-understanding boundary, making translation useful for solving it. We use $\delta=0.5$ throughout our experiments, so that $u(q)=1$ reflects a substantial empirical benefit from translation.

\subsection{\textsc{Luar}: Two-Stage Training}
\label{sec:luar}

Given this boundary estimate, \textsc{Luar} trains the model in two stages. Stage I extends the model's action space by enabling it to produce translator-augmented reasoning traces. Stage II then optimizes this behavior with reinforcement learning, so that the model learns when translation should be invoked.
\sparagraph{Stage I: Translator-Call SFT.}
\label{sec:sft}
Stage I performs supervised fine-tuning (SFT) as a warm-up step before reinforcement learning. We sample responses from the base model under both direct and translator-augmented reasoning modes, and construct supervised targets from correct samples. For each query $q$, let $\mathcal{Y}_{d}(q)$ and $\mathcal{Y}_{t}(q)$ denote the sets of correct responses sampled under the two modes, respectively, where translator-augmented responses use the translator-call prefix in Figure~\ref{fig:translator-prefix}.
We select the SFT mode $m(q) \in \{d,t\}$ as
\begin{equation}
m(q) =
\begin{cases}
t, & \text{if } t_{\mathrm{acc}}(q) > d_{\mathrm{acc}}(q), \\
d, & \text{otherwise}.
\end{cases}
\end{equation}
Since Stage I encourages exploration rather than defining the final policy, we use this softer rule instead of the stricter translation-usefulness label $u(q)$ in~\eqref{eq:boundary-label}.
This introduces potentially useful translator-augmented trajectories within the model's sampling space, allowing Stage II to learn when to invoke translation through reinforcement learning.
We then sample one correct response from the selected mode:

\begin{equation}
y_{\mathrm{sft}}(q) \sim
\mathrm{Uniform}\left(\mathcal{Y}_{m(q)}(q)\right),
\end{equation}
removing queries for which the selected mode has no correct response.
Finally, we fine-tune the model to predict $y_{\mathrm{sft}}(q)$ given $q$.
\sparagraph{Stage II: Boundary-Aware GRPO.}
\label{sec:rl}
After Stage I, the model can generate both direct and translator-augmented responses. Stage II further optimizes this behavior through online reinforcement learning, so that the model learns to invoke translation selectively. To this end, we use GRPO~\cite{shao2024deepseekmathpushinglimitsmathematical} with a boundary-aware reward (Eq.~\eqref{eq:boundary_reward}) that jointly scores task correctness and translator-call decisions. For each query $q$, we sample a group of $G$ responses $\{y_i\}_{i=1}^{G}$ from the old policy $\pi_{\theta_{\mathrm{old}}}$. Unlike the boundary estimation in \S\ref{sec:boundary-estimation}, this stage does not force a balanced number of direct and translator-augmented responses. Each response is generated by the model's current policy, which may decide whether to call the translator within its reasoning trace. When the model calls the translator, we provide the \textbf{reference} English translation of the input query as the translator response during training. This lets the model learn when translation is useful without confounding effects from translation quality. For each response $y_i$, let $c(y_i)\in\{0,1\}$ indicate whether the response gives the correct answer and conforms to the required output format,\footnote{See Appendix~\ref{app:training-details} for details on the format-checking criteria.} and let $t(y_i)\in\{0,1\}$ indicate whether it calls the translator. Using the translation-usefulness label $u(q)$ from~\eqref{eq:boundary-label}, we define the \textbf{boundary-aware reward} as
\begin{equation}
\label{eq:boundary_reward}
\resizebox{0.88\columnwidth}{!}{$
r(q,y_i)=
\begin{cases}
1+\gamma, & c(y_i)=1,\; t(y_i)=1,\; u(q)=1,\\
1, & c(y_i)=1,\; t(y_i)=0,\\
1-\gamma, & c(y_i)=1,\; t(y_i)=1,\; u(q)=0,\\
0, & c(y_i)=0,
\end{cases}
$}
\end{equation}
where $\gamma=0.1$ in our experiments. This reward prioritizes task correctness while encouraging translator calls only when translation is estimated to be useful. Among correct responses, direct reasoning receives the base reward, while translator calls receive a bonus when $u(q)=1$ and a penalty when $u(q)=0$.\footnote{When all responses in the sampled group use the same mode, we omit the translator-use adjustment $\gamma$ and set $r(q,y_i)=c(y_i)$.}
We normalize the rewards within each sampled group to obtain group-relative advantages:
\begin{equation}
\hat{A}_i =
\frac{
r(q,y_i)-\mathrm{mean}(\{r(q,y_j)\}_{j=1}^{G})
}{
\mathrm{std}(\{r(q,y_j)\}_{j=1}^{G})
}.
\end{equation}
We then optimize the clipped token-level objective
\begin{equation}
\resizebox{0.88\columnwidth}{!}{$
\begin{aligned}
\mathcal{J}_{\mathrm{GRPO}}(\theta)
&=
\mathbb{E}_{\{y_i\}_{i=1}^{G} \sim \pi_{\theta_{\mathrm{old}}}(\cdot \mid q)}
\Bigg[
\frac{1}{\sum_{i=1}^{G}|y_i|}
\sum_{i=1}^{G}\sum_{t=1}^{|y_i|} \\
&\quad
\min\Big(
\rho_{i,t}\hat{A}_{i},
\mathrm{clip}(\rho_{i,t},1-\epsilon_l,1+\epsilon_h)\hat{A}_{i}
\Big)
\Bigg],
\end{aligned}
$}
\end{equation}
where $\rho_{i,t}$ is the token-level importance sampling ratio.
Following DAPO-style training~\cite{yu2025dapoopensourcellmreinforcement}, we minimize the token-level policy gradient loss with $\epsilon_l=0.2$ and $\epsilon_h=0.3$, without a KL divergence loss. For both Stages I and II, we mask out the tokens between \texttt{<translator\_response>} and \texttt{</translator\_response>} during loss computation for training stability, following \citet{feng2025retoolreinforcementlearningstrategic}.

%% file: latex/figure/main_figure.tex
\begin{figure*}[!t]
\centering
\includegraphics[width=\textwidth]{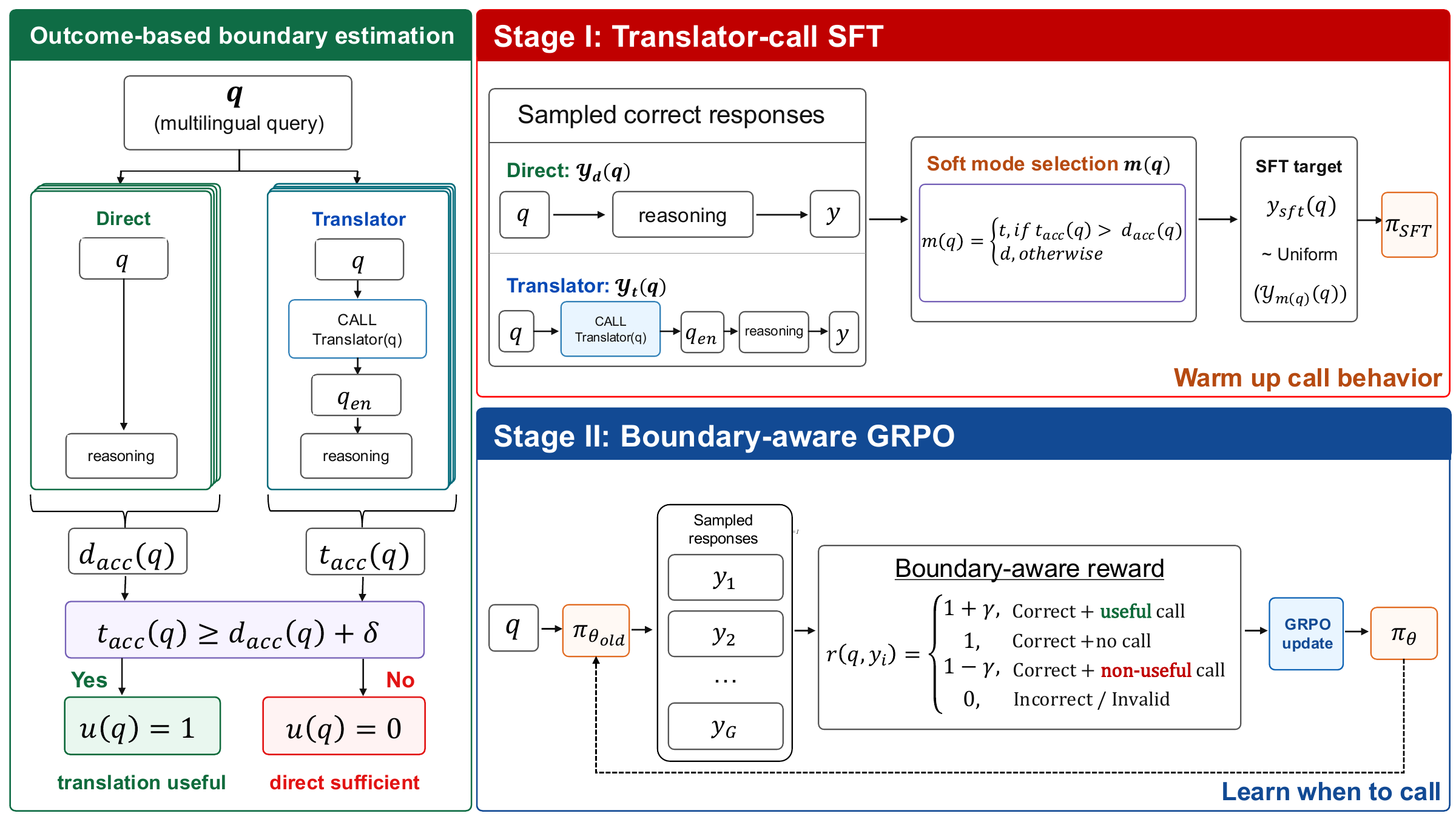}
\caption{Overview of \textsc{Luar}. We first derive an outcome-based translation-usefulness label $u(q)$ by comparing direct and translator-augmented reasoning outcomes. Stage I uses supervised fine-tuning to warm up both reasoning trajectories, and Stage II applies GRPO with a boundary-aware reward to learn when to invoke translation.}
\label{fig:luar-overview}

\end{figure*}

%% file: latex/content/4_experiments.tex
\section{Experiments}

\subsection{Experimental Setup}
\label{sec:experimental-setup}

\input{latex/table/table1_mode_selection_macro_f1}

\sparagraph{Models.}
We primarily evaluate \textsc{Luar} on Qwen3-4B and additionally on Qwen3-8B~\cite{yang2025qwen3technicalreport}.

\begin{figure*}[t]
    \centering
    \includegraphics[width=\linewidth]{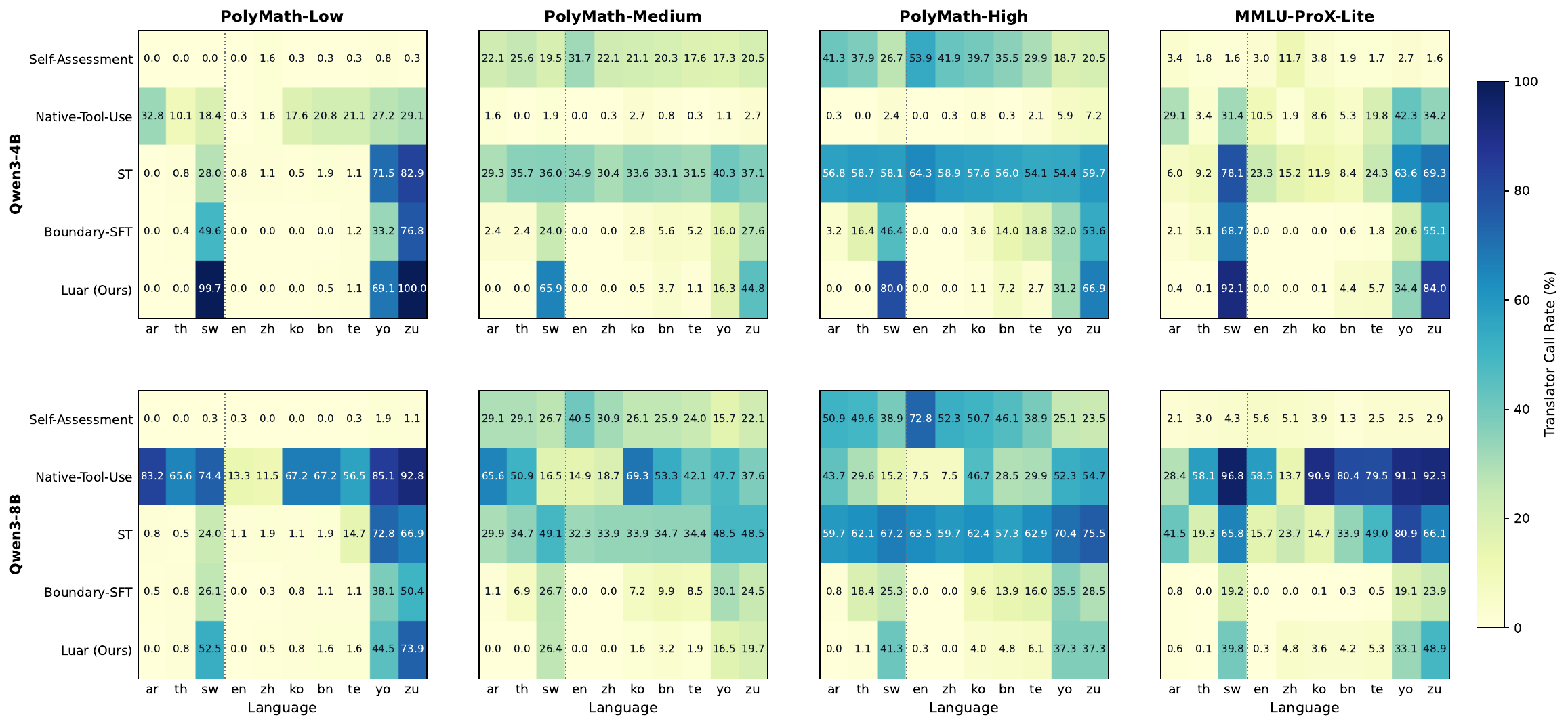}
    \caption{Translator-call rates across languages and datasets. \textsc{Luar} concentrates translator calls on low-resource languages and varies its call rate across datasets, whereas \textsc{ST} frequently calls the translator for high-resource languages on harder PolyMath splits.}
    \label{fig:translator_call_rate}
\end{figure*}

\sparagraph{Datasets.}
We evaluate on two multilingual reasoning benchmarks. PolyMath~\cite{wang2025polymath} measures mathematical reasoning, and we use its low-, medium-, and high-difficulty subsets, each with 125 test examples per language. To evaluate out-of-domain generalization beyond mathematics, we also use STEM-related categories from MMLU-ProX-Lite~\cite{xuanmmluprox}, which contains 257 test examples per language. Evaluation covers 10 typologically diverse languages. Following~\citet{joshi-etal-2020-state}, we group them into high-resource languages (English (en), Chinese (zh), Arabic (ar), Korean (ko)), mid-resource languages (Thai (th), Bengali (bn)), and low-resource languages (Swahili (sw), Telugu (te), Yoruba (yo), Zulu (zu)). For training, we use the training split of DeepScaleR~\cite{deepscaler2025}, which is an English mathematical problem-solving dataset. After contamination filtering against PolyMath, we translate the remaining questions into Arabic, Thai, and Swahili using GPT-4.1~\cite{openai2025gpt4.1}. We choose these languages to expose the policy to different scripts and resource conditions. Stage I uses 10K examples, and Stage II uses a disjoint set of 10K examples. In both stages, examples are evenly distributed across the three training languages.
Appendix~\ref{app:dataset-training-data} provides further details on the datasets and training-data construction.

\sparagraph{Evaluation metrics.}
We report task \textbf{accuracy} and mode-selection \textbf{macro-F1}. The latter compares translator-call decisions with the outcome-based usefulness label $u(q)$ defined in \S\ref{sec:boundary-estimation}. We estimate evaluation labels separately for each method using its final evaluated checkpoint: the selected fine-tuned model for trained methods and the base model otherwise. Macro-F1 therefore measures routing relative to each method's own estimated translation benefit. Appendix~\ref{app:evaluation-details} provides details on label construction.

\sparagraph{Baselines.}
We compare against four groups of baselines. The first three groups test alternative ways of making translator-call decisions, while the last group tests whether training alone can close the accuracy gap without using an external translator at inference time. \textbf{(1) Prompt-based methods:} \textsc{Self-Assessment} prompts the model to solve the problem and then report whether the original input was understandable. If it reports understanding failure, inference is rerun with translation. 
\textsc{Native-Tool-Use} equips the model with a translator function through its default function-calling interface, allowing the model to decide whether to call the translator when the input is difficult to interpret. \textbf{(2) External detector-based methods:} 
Selective Translation (\textsc{ST})~\citep{kang2026multilingualreasoninggapsemerge} first lets the model generate a complete reasoning trace. A prober trained to detect understanding failures in the model then identifies queries to rerun with translation. \textbf{(3) SFT-based method:}
\textsc{Boundary-SFT} internalizes the same outcome-based boundary signal (\S\ref{sec:boundary-estimation}) through supervised fine-tuning. For each query, it fine-tunes on correct direct or translator-augmented traces selected according to the \textit{strict} translation-usefulness label $u(q)$.
\textbf{(4) Translator-free training methods:} We compare with standard \textsc{GRPO}~\cite{shao2024deepseekmathpushinglimitsmathematical}, as well as \textsc{Q-Align}~\cite{zhu-etal-2024-question} and \textsc{TAPO}~\cite{huang2026tapo}, which we adapt for RLMs. We include these baselines in the accuracy comparison to cover three training-only alternatives: \textsc{Q-Align} improves language understanding through translation-aligned supervision, standard \textsc{GRPO} improves reasoning using the same RL optimization setup as \textsc{Luar} but with the correctness reward $c(y)$, and \textsc{TAPO} jointly trains the model's internal language understanding and reasoning through translation-augmented policy optimization. We also report \textsc{Full-Translation} as a high-cost reference that translates every input. \textsc{Base} denotes direct inference with the original Qwen checkpoint on the original-language queries. All methods using an external translator, including \textsc{Luar}, use GPT-4.1 at evaluation time. Further details are provided in Appendix~\ref{app:baseline-details}.

\sparagraph{Implementation details.}
We implement \textsc{Luar} using VerlTool~\cite{jiang2025verltool}. Stage I uses supervised fine-tuning for 3 epochs with a learning rate of $1\times10^{-5}$ and batch size 16. Stage II uses reinforcement learning for 150 steps with a learning rate of $2\times10^{-6}$, batch size 32, mini-batch size 16, group size 8, and rollout temperature 1.0. We use the same SFT and RL configurations for all corresponding training baselines. For evaluation, we average results over three random seeds and use the same decoding configuration across all methods. Appendix~\ref{app:training-details} and Appendix~\ref{app:evaluation-details} provide additional training and evaluation details.

\input{latex/table/table2_accuracy}

\input{latex/table/table3_per_language_accuracy}

\input{latex/table/table4_efficiency}

\subsection{Main Results}
\textbf{\textsc{Luar} effectively learns to invoke translation when it benefits reasoning.} We first test whether \textsc{Luar} learns an effective mode-selection policy from the outcome-based boundary signal. Table~\ref{tab:table1_mode_selection_macro_f1} reports mode-selection macro-F1 relative to the estimated boundary labels. Overall, \textsc{Luar} achieves the best average macro-F1 for both models, indicating that it effectively learns to invoke translation when it benefits reasoning.\footnote{Appendix~\ref{app:threshold_robustness} shows that \textsc{Luar} maintains strong mode-selection performance when evaluated under different boundary thresholds $\delta$.} Prompt-based methods and \textsc{Boundary-SFT} perform worse, suggesting that prompting or SFT alone is less effective for boundary-aware decisions. Compared with \textsc{ST}, \textsc{Luar} achieves higher macro-F1 in all evaluated settings except PolyMath-Low and PolyMath-Medium on Qwen3-8B. To better understand the translator-call behavior underlying these results, we next compare call rates across languages and datasets.

\sparagraph{\textsc{Luar} learns a robust and targeted translator-call policy.}
Figure~\ref{fig:translator_call_rate} shows that \textsc{Luar} rarely invokes translation for high-resource languages and concentrates calls on low-resource languages, including unseen Yoruba and Zulu. In contrast, \textsc{ST} calls the translator even for high-resource languages, including 63.5\% of English examples on PolyMath-High with Qwen3-8B. To test whether \textsc{Luar}'s calls depend on more than language identity, we compare queries within Yoruba and Zulu (Appendix~\ref{app:within-language-routing}). In both languages, it calls the translator more often for queries the base model consistently solves only with translation than for those consistently solved directly (61.3\% vs.\ 22.8\% in Yoruba and 94.3\% vs.\ 59.0\% in Zulu). These results suggest that \textsc{Luar}'s calls reflect query-level translation usefulness beyond language identity.

\subsection{Task Accuracy}

\sparagraph{\textsc{Luar} improves multilingual reasoning performance.}
Table~\ref{tab:table2_accuracy_qwen3_4b} reports Qwen3-4B final-answer accuracy and translator use. \textsc{Full-Translation} serves as a high-cost reference. Among methods that do not translate every input, \textsc{Luar} achieves the highest average accuracy, improving \textsc{Base} from 44.93 to 54.39 while translating 20.3\% of examples. \textsc{Luar}'s gains are especially strong on PolyMath, where it achieves the best accuracy among these methods across all three difficulty levels. On MMLU-ProX-Lite, it improves on \textsc{Base} but trails \textsc{ST}, whose broader translator use includes unnecessary calls on high-resource languages (Figure~\ref{fig:translator_call_rate}). Translator-free training baselines yield smaller gains and remain below methods that invoke translation, supporting the value of external translation.

\sparagraph{\textsc{Luar}'s accuracy gains are concentrated in languages where direct reasoning is weakest.} Table~\ref{tab:per_language_accuracy_polymath_low} shows per-language accuracy on PolyMath-Low. The base Qwen3-4B model performs much worse on low-resource languages. \textsc{Q-Align} and \textsc{TAPO} improve accuracy on Swahili, a seen low-resource language, but transfer poorly to unseen low-resource languages such as Yoruba and Zulu. By contrast, \textsc{Luar} reaches 56.27 on Yoruba and 87.20 on Zulu and has the highest average among methods that do not translate every input. These gains show that its translator-use policy transfers to languages absent from task-specific training, enabling the model to benefit from external translation. Additional per-language accuracy results are reported in Appendix~\ref{app:per-language-results}.

\subsection{Efficiency}

We evaluate \textsc{Luar}'s inference efficiency by comparing inference cost and latency on Qwen3-4B. Table~\ref{tab:efficiency_overall_qwen3_4b} shows that \textsc{Full-Translation} incurs 2.02$\times$ the estimated cost of \textsc{Base}. In contrast, \textsc{Luar} achieves average accuracy comparable to that of \textsc{Full-Translation} (Table~\ref{tab:table2_accuracy_qwen3_4b}) at approximately 46\% of its inference cost by concentrating translator calls on queries where translation benefits reasoning.\footnote{Shorter reasoning traces also contribute to lower costs and latency. Even with \textsc{Luar}'s mean reasoning output length matched to \textsc{Base}'s, its estimated total cost remains lower than those of \textsc{ST} and \textsc{Full-Translation} (Appendix~\ref{app:inference_efficiency_measurement}).} \textsc{Luar} also achieves lower latency than \textsc{ST} (9.32 vs.\ 14.09 seconds). It makes translation decisions and continues reasoning within a single trajectory, avoiding \textsc{ST}'s rerun of selected queries after direct reasoning. These results highlight \textsc{Luar}'s efficiency gains from selectively invoking the translator and internalizing translation decisions within the reasoning policy.

\begin{figure}[t]
\centering

\includegraphics[width=0.85\columnwidth]{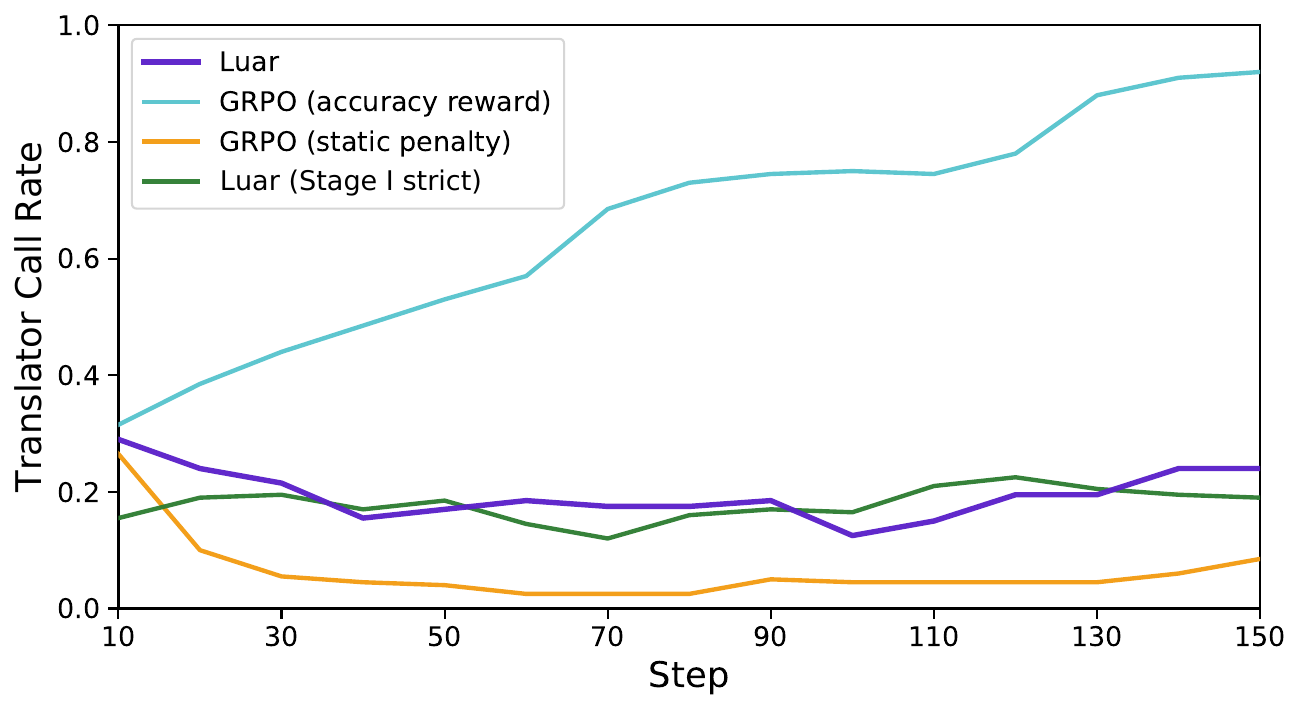}
\caption{Validation translator-call rates during training.}
\label{fig:training_dynamics}
\end{figure}
\input{latex/table/table5_ablation}

\subsection{Ablation Study} 
We compare \textsc{Luar} with four variants to analyze how training shapes translator-call behavior. First, \textit{Stage I only} omits Stage II. Second, \textit{GRPO with accuracy reward} uses only the task-success reward, $r=c(y)$, where $c(y)$ denotes the answer-correct and format-correct indicator used in Eq.~\ref{eq:boundary_reward}. Third, \textit{GRPO with static penalty} subtracts a fixed $\gamma=0.1$ penalty from successful translator-call responses, regardless of whether translation is useful. Finally, \textsc{Luar} (Stage I strict) uses the strict usefulness label for Stage I target selection, instead of the softer rule $t_{\mathrm{acc}}(q)>d_{\mathrm{acc}}(q)$ used by \textsc{Luar}.

\sparagraph{\textsc{Luar}'s boundary-aware reward is effective for selective translator calling.}
Starting from the same Stage I model, \textsc{Luar} improves average accuracy from 53.14 to 54.39 and mode-selection macro-F1 from 70.79 to 78.26 relative to GRPO with the accuracy reward (Table~\ref{tab:ablation_table_macro_f1}). Figure~\ref{fig:training_dynamics} shows that training with only the correctness reward increasingly favors translation, whereas the static penalty suppresses calls. \textsc{Luar} maintains selective use by conditioning the reward adjustment on estimated translation usefulness. Beyond the RL reward, the Stage I target-selection rule also affects performance. The strict Stage I variant underperforms \textsc{Luar}, suggesting that softer target selection helps expose translator-call behavior before refinement with the boundary-aware reward.

%% file: latex/table/table1_mode_selection_macro_f1.tex
\begin{table*}[!t]
\centering
\scriptsize
\setlength{\tabcolsep}{3pt}
\resizebox{\textwidth}{!}{%
\begin{tabular}{lccccc}
\toprule
\textbf{Method} & \textbf{PolyMath-Low} & \textbf{PolyMath-Medium} & \textbf{PolyMath-High} & \textbf{MMLU-ProX-Lite} & \textbf{Average (↑)} \\
\midrule
\multicolumn{6}{l}{\textbf{\textit{Qwen3-4B}}} \\
\textsc{Self-Assessment} & $41.10_{\pm 0.28}$ & $56.44_{\pm 2.55}$ & $52.93_{\pm 1.48}$ & $47.83_{\pm 1.04}$ & $49.57_{\pm 0.83}$ \\
\textsc{Native-Tool-Use} & $57.60_{\pm 1.90}$ & $49.61_{\pm 0.19}$ & $48.54_{\pm 2.36}$ & $60.03_{\pm 0.48}$ & $53.95_{\pm 1.13}$ \\
\textsc{ST} & $79.26_{\pm 0.50}$ & $64.59_{\pm 0.42}$ & $58.24_{\pm 1.89}$ & $65.44_{\pm 0.13}$ & $66.88_{\pm 0.52}$ \\
\midrule
\textsc{Boundary-SFT} & $73.14_{\pm 0.69}$ & $62.22_{\pm 4.75}$ & $66.85_{\pm 1.32}$ & $65.21_{\pm 0.44}$ & $66.85_{\pm 1.80}$ \\
\midrule
\rowcolor{gray!12} \textbf{\textsc{Luar}} & $\mathbf{87.64}_{\pm 0.33}$ & $\mathbf{75.22}_{\pm 1.84}$ & $\mathbf{73.74}_{\pm 1.11}$ & $\mathbf{76.45}_{\pm 1.02}$ & $\mathbf{78.26}_{\pm 0.42}$ \\
\midrule
\multicolumn{6}{l}{\textbf{\textit{Qwen3-8B}}} \\
\textsc{Self-Assessment} & $43.48_{\pm 0.83}$ & $54.28_{\pm 2.56}$ & $48.60_{\pm 1.92}$ & $48.04_{\pm 0.38}$ & $48.60_{\pm 0.47}$ \\
\textsc{Native-Tool-Use} & $58.41_{\pm 3.53}$ & $39.23_{\pm 0.35}$ & $45.60_{\pm 2.03}$ & $41.36_{\pm 0.59}$ & $46.15_{\pm 0.67}$ \\
\textsc{ST} & $\mathbf{82.92}_{\pm 0.94}$ & $\mathbf{69.19}_{\pm 0.33}$ & $60.73_{\pm 0.84}$ & $61.13_{\pm 0.05}$ & $68.49_{\pm 0.11}$ \\
\midrule
\textsc{Boundary-SFT} & $69.20_{\pm 0.44}$ & $65.90_{\pm 1.72}$ & $60.54_{\pm 1.32}$ & $59.47_{\pm 0.50}$ & $63.78_{\pm 0.55}$ \\
\midrule
\rowcolor{gray!12} \textbf{\textsc{Luar}} & $78.32_{\pm 0.11}$ & $65.61_{\pm 1.29}$ & $\mathbf{69.74}_{\pm 1.34}$ & $\mathbf{68.18}_{\pm 0.55}$ & $\mathbf{70.46}_{\pm 0.44}$ \\
\bottomrule
\end{tabular}%
}
\caption{Mode-selection macro-F1 (\%) for translator-call decisions. Numbers are mean$_{\pm\mathrm{std}}$ over seeds.}
\label{tab:table1_mode_selection_macro_f1}

\end{table*}

%% file: latex/table/table2_accuracy.tex
\begin{table*}[!t]
\centering
\scriptsize
\setlength{\tabcolsep}{3pt}
\resizebox{\textwidth}{!}{%
\begin{tabular}{lccccl}
\toprule
\textbf{Method} & \textbf{PolyMath-Low} & \textbf{PolyMath-Medium} & \textbf{PolyMath-High} & \textbf{MMLU-ProX-Lite} & \textbf{Average (Use \%)} \\
\midrule
\textsc{Base} & $64.40_{\pm 0.16}$ & $32.05_{\pm 0.24}$ & $19.76_{\pm 0.08}$ & $63.49_{\pm 0.49}$ & $44.93_{\pm 0.14}$ (0.0\%) \\
\midrule
\textsc{Full-Translation} & $88.59_{\pm 0.44}$ & $35.79_{\pm 1.09}$ & $24.24_{\pm 0.79}$ & $72.94_{\pm 0.31}$ & $55.39_{\pm 0.05}$ (100.0\%) \\
\specialrule{0.10em}{0.05em}{0.05em}
\textsc{Self-Assessment} & $63.52_{\pm 0.76}$ & $36.00_{\pm 0.29}$ & $23.95_{\pm 0.17}$ & $64.36_{\pm 0.34}$ & $46.96_{\pm 0.20}$ (15.0\%) \\
\textsc{Native-Tool-Use} & $70.27_{\pm 0.65}$ & $34.93_{\pm 0.36}$ & $23.73_{\pm 0.23}$ & $65.77_{\pm 0.35}$ & $48.68_{\pm 0.16}$ (9.9\%) \\
\textsc{ST} & $79.36_{\pm 0.16}$ & $35.47_{\pm 0.54}$ & $24.32_{\pm 0.48}$ & $\mathbf{71.57}_{\pm 0.31}$ & $52.68_{\pm 0.31}$ (35.5\%) \\
\textsc{Boundary-SFT} & $74.20_{\pm 1.53}$ & $31.56_{\pm 0.96}$ & $20.68_{\pm 0.40}$ & $63.60_{\pm 0.36}$ & $47.51_{\pm 0.05}$ (14.7\%) \\
\midrule
\textsc{Q-Align} & $62.43_{\pm 0.28}$ & $30.64_{\pm 0.40}$ & $19.07_{\pm 0.65}$ & $59.79_{\pm 0.25}$ & $42.98_{\pm 0.12}$ (0.0\%) \\
\textsc{GRPO} & $63.71_{\pm 0.88}$ & $35.79_{\pm 0.60}$ & $22.96_{\pm 0.65}$ & $61.41_{\pm 0.20}$ & $45.97_{\pm 0.16}$ (0.0\%) \\
\textsc{TAPO} & $61.20_{\pm 0.21}$ & $36.13_{\pm 0.18}$ & $23.41_{\pm 0.68}$ & $61.25_{\pm 0.12}$ & $45.50_{\pm 0.19}$ (0.0\%) \\
\midrule
\rowcolor{gray!12} \textbf{\textsc{Luar}} & $\mathbf{82.88}_{\pm 0.65}$ & $\mathbf{37.95}_{\pm 1.19}$ & $\mathbf{28.32}_{\pm 0.63}$ & $68.43_{\pm 0.34}$ & $\mathbf{54.39}_{\pm 0.07}$ (20.3\%) \\
\bottomrule
\end{tabular}%
}
\caption{Qwen3-4B accuracy (\%; mean$_{\pm\mathrm{std}}$ over seeds). Parentheses in the Average column show translator use (\%). Bold marks the best method that does not translate every input.}
\label{tab:table2_accuracy_qwen3_4b}

\end{table*}

%% file: latex/table/table3_per_language_accuracy.tex
\begin{table*}[t]
\centering
\small
\setlength{\tabcolsep}{4.5pt}
\resizebox{\textwidth}{!}{%
\begin{tabular}{lccccccccccc}
\toprule
\textbf{Method} 
& \multicolumn{3}{c}{\textbf{Seen}} 
& \multicolumn{7}{c}{\textbf{Unseen}} 
& \textbf{Avg.} \\
\cmidrule(lr){2-4}\cmidrule(lr){5-11}
& \textbf{ar} & \textbf{th} & \textbf{sw}
& \textbf{en} & \textbf{zh} & \textbf{ko} & \textbf{bn} & \textbf{te} & \textbf{yo} & \textbf{zu}
& \\
\midrule
\textsc{Base} & 89.33 & 83.73 & 32.53 & 95.73 & 86.93 & 89.60 & 82.93 & 66.67 & 8.53 & 8.00 & 64.40 \\
\midrule
\textsc{Full-Translation} & \cellcolor{green!3}91.20 & \cellcolor{green!12}90.40 & \cellcolor{green!35}84.80 & 95.73 & \cellcolor{green!10}92.80 & \cellcolor{green!2}90.93 & \cellcolor{green!14}91.20 & \cellcolor{green!27}81.87 & \cellcolor{green!35}78.40 & \cellcolor{green!35}88.53 & \cellcolor{green!35}88.59 \\
\specialrule{0.10em}{0.05em}{0.05em}
\textsc{Self-Assessment} & \cellcolor{green!2}90.67 & \cellcolor{red!2}82.40 & \cellcolor{red!3}30.67 & \cellcolor{red!1}94.93 & \cellcolor{green!2}88.00 & \cellcolor{red!1}89.33 & \cellcolor{red!1}82.67 & \cellcolor{green!1}66.93 & \cellcolor{red!6}5.07 & \cellcolor{red!6}4.53 & \cellcolor{red!2}63.52  \\
\textsc{Native-Tool-Use} & \cellcolor{green!1}89.87 & \cellcolor{red!1}83.47 & \cellcolor{green!14}40.53 & \cellcolor{red!3}94.13 & \cellcolor{green!3}88.80 & 89.60 & \cellcolor{green!1}83.73 & \cellcolor{green!6}70.13 & \cellcolor{green!35}29.33 & \cellcolor{green!35}33.07 & \cellcolor{green!10}70.27  \\
\textsc{ST} & \cellcolor{green!1}90.13 & \cellcolor{green!3}85.33 & \cellcolor{green!35}52.53 & \cellcolor{green!1}96.00 & \cellcolor{green!2}88.27 & \cellcolor{green!1}89.87 & \cellcolor{green!2}84.27 & \cellcolor{green!6}69.87 & \cellcolor{green!35}58.93 & \cellcolor{green!35}78.40 & \cellcolor{green!26}79.36 \\
\textsc{Boundary-SFT} & \cellcolor{green!1}90.00 & \cellcolor{red!1}83.20 & \cellcolor{green!35}54.80 & \cellcolor{red!2}94.80 & \cellcolor{red!6}83.60 & \cellcolor{red!9}84.40 & \cellcolor{red!6}79.60 & \cellcolor{green!5}69.60 & \cellcolor{green!35}30.80 & \cellcolor{green!35}71.20 & \cellcolor{green!17}74.20 \\
\midrule
\textsc{Q-Align} & \cellcolor{red!3}87.73 & \cellcolor{green!1}84.00 & \cellcolor{green!10}38.40 & 95.73 & 86.93 & \cellcolor{red!10}84.00 & \cellcolor{red!10}77.33 & \cellcolor{red!6}63.20 & \cellcolor{red!10}2.67 & \cellcolor{red!7}4.27 & \cellcolor{red!3}62.43 \\
\textsc{GRPO} & \cellcolor{red!2}88.00 & \cellcolor{red!1}83.20 & \cellcolor{red!5}29.60 & \cellcolor{red!1}94.93 & \cellcolor{green!1}87.73 & \cellcolor{red!1}89.33 & \cellcolor{red!4}80.80 & \cellcolor{red!2}65.60 & \cellcolor{green!5}11.20 & \cellcolor{red!2}6.67 & \cellcolor{red!1}63.71 \\
\textsc{TAPO} & \cellcolor{red!11}83.20 & \cellcolor{red!5}81.07 & \cellcolor{green!12}39.20 & \cellcolor{red!1}95.47 & \cellcolor{red!2}85.87 & \cellcolor{red!9}84.53 & \cellcolor{red!19}72.00 & \cellcolor{red!9}61.33 & \cellcolor{red!8}3.73 & \cellcolor{red!4}5.60 & \cellcolor{red!6}61.20 \\
\midrule
\rowcolor{gray!12} \textsc{\textbf{Luar}} & \cellcolor{green!1}89.60 & \cellcolor{green!2}85.07 & \cellcolor{green!35}89.60 & 95.73 & \cellcolor{green!4}89.07 & \cellcolor{red!5}86.67 & \cellcolor{red!2}81.60 & \cellcolor{green!2}68.00 & \cellcolor{green!35}56.27 & \cellcolor{green!35}87.20 & \cellcolor{green!32}\textbf{82.88} \\
\bottomrule
\end{tabular}
}
\caption{PolyMath-Low accuracy by language on Qwen3-4B (\%; mean over seeds). Bold marks the best average without translating every input. Colors show changes from \textsc{Base}, clipped at $\pm$20 points.}\label{tab:per_language_accuracy_polymath_low}
\end{table*}

%% file: latex/table/table4_efficiency.tex
\begin{table*}[!t]
\centering
\small
\setlength{\tabcolsep}{7pt}
\begin{tabular}{@{}lrrrrr@{}}
\toprule
& \textbf{Latency} & \multicolumn{3}{c}{\textbf{Cost (USD)}} & \textbf{Translator} \\
\cmidrule(lr){3-5}
\textbf{Method} & \textbf{(s)} & \textbf{Total} & \textbf{Reasoning} & \textbf{Translation} & \textbf{use (\%)} \\
\midrule
\textsc{Base} & 11.77 & \$0.00191 (1.00$\times$) & \$0.00191 & \$0.00000 & 0.0 \\
\textsc{ST} & 14.09 & \$0.00300 (1.57$\times$) & \$0.00227 & \$0.00073 & 30.0 \\
\textsc{Full-Translation} & 11.97 & \$0.00385 (2.02$\times$) & \$0.00200 & \$0.00184 & 100.0 \\
\rowcolor{gray!12}\textbf{\textsc{Luar}} & 9.32 & \$0.00178 (0.93$\times$) & \$0.00142 & \$0.00036 & 20.5 \\
\bottomrule
\end{tabular}
\caption{Qwen3-4B inference cost, latency, and translator use, averaged over all queries in one evaluation run. Costs are USD per query, including reruns and translation instructions. Reasoning and translation costs count Qwen3-4B and GPT-4.1 input and output tokens, respectively. Parentheses give total cost relative to \textsc{Base}. See Appendix~\ref{app:inference_efficiency_measurement}.}
\label{tab:efficiency_overall_qwen3_4b}
\end{table*}

%% file: latex/table/table5_ablation.tex
\begin{table}[!t]
\centering
\small
\setlength{\tabcolsep}{6pt}
\resizebox{\columnwidth}{!}{%
\begin{tabular}{lrr}
\toprule
\textbf{Method} & \textbf{Accuracy (\%)} & \textbf{Macro-F1 (\%)} \\
\midrule
Stage I only & $50.46_{\pm 0.16}$ & $71.85_{\pm 0.39}$ \\
+ GRPO, accuracy reward & $53.14_{\pm 0.08}$ & $70.79_{\pm 0.79}$ \\
+ GRPO, static penalty ($\gamma=0.1$) & $50.33_{\pm 0.12}$ & $70.55_{\pm 0.17}$ \\
\rowcolor{gray!12} \textbf{\textsc{Luar} (Ours)} & $\mathbf{54.39}_{\pm 0.07}$ & $\mathbf{78.26}_{\pm 0.42}$ \\
\midrule
\textsc{Luar} (Stage I strict) & $53.66_{\pm 0.06}$ & $74.86_{\pm 0.59}$ \\
\bottomrule
\end{tabular}%
}
\caption{Effect of RL training and reward designs on Qwen3-4B. Values are averaged over datasets and seeds.}
\label{tab:ablation_table_macro_f1}

\end{table}

%% file: latex/content/5_conclusion.tex
\section{Conclusion}
We introduced \textsc{Luar}, a Language Understanding Boundary-aware Reinforcement Learning framework for multilingual reasoning. \textsc{Luar} improves reasoning accuracy across multilingual benchmarks by selectively invoking translation when direct understanding is unreliable. The learned translator-use policy also yields substantial accuracy gains on low-resource languages absent from task-specific training, including Yoruba and Zulu. Future work could extend this policy to other multilingual tasks and abstention.

%% file: latex/content/limitations.tex
\section*{Limitations}

Our evaluation includes typologically diverse languages across different resource conditions and covers both mathematical and STEM reasoning benchmarks, which differ in domain and question format. This provides evidence of \textsc{Luar}'s effectiveness across varied multilingual reasoning settings. However, the evaluated languages do not represent the full diversity of language families, and multilingual reasoning encompasses tasks beyond mathematics and STEM, including commonsense and knowledge-intensive reasoning. Extending the evaluation to a broader range of languages and tasks would provide a more comprehensive understanding of the generality of boundary-aware translation-use policies. Although Qwen3-4B and Qwen3-8B are widely used reasoning language models, computational constraints limit our training experiments to two scales within the same model family. Further evaluation across different model families and scales would help assess the broader generalizability of \textsc{Luar}'s effectiveness.

To keep translation-usefulness estimation computationally feasible, we use eight sampled responses per query, with four responses from each reasoning mode. This allows us to estimate the relative benefit of translation for individual queries, although the resulting labels may be sensitive to sampling variability. Increasing the number of samples would help assess the stability of the estimated boundaries and the robustness of our mode-selection evaluation.

%% file: latex/content/ethical_considerations.tex
\section*{Ethical Considerations}
In this work, we use DeepScaleR~\cite{deepscaler2025} for training, and PolyMath~\cite{wang2025polymath} and MMLU-ProX-Lite~\cite{xuanmmluprox} for evaluation. These datasets are licensed under MIT, Apache 2.0, and MIT, respectively. We conduct experiments with Qwen3-4B/8B~\cite{yang2025qwen3technicalreport} and gpt-oss-20b/120b~\cite{agarwal2025gpt}, all of which are licensed under Apache 2.0. All datasets and models are used strictly for research purposes and in accordance with their intended use. We used ChatGPT and Codex for language polishing, grammar correction, and programming assistance. All ideas, contributions, and research decisions are solely those of the authors.

%% file: latex/content/acknowledgement.tex
\section*{Acknowledgments}
This research was supported by Culture, Sports and Tourism R\&D Program through the Korea Creative Content Agency grant funded by the Ministry of Culture, Sports and Tourism in 2025 (Project Name: Development of an AI-Based Korean Diagnostic System for Efficient Korean Speaking Learning by Foreigners, Project Number: RS-2025-02413038, Contribution Rate: 45\%); by the IITP(Institute of Information \& Coummunications Technology Planning \& Evaluation)-ITRC(Information Technology Research Center) grant funded by the Korea government(Ministry of Science and ICT)(IITP-2026-RS-2024-00437866, Contribution Rate: 45\%); and by the Institute of Information \& communications Technology Planning \& Evaluation (IITP) grant funded by the Korea government(MSIT) (No.RS-2019-II191906, Artificial Intelligence Graduate School Program(POSTECH), Contribution Rate: 10\%).

%% file: latex/appendix/appendix_section_A.tex
\begin{table*}[!t]
\centering
\scriptsize
\setlength{\tabcolsep}{5pt}
\renewcommand{\arraystretch}{1.18}
\begin{tabular}{>{\raggedright\arraybackslash}p{0.18\textwidth}|>{\raggedright\arraybackslash}p{0.76\textwidth}}
\toprule
\textbf{Swahili question} &
{Dada wa Doubtfire wanaendesha wakielekea nyumbani na watoto 7 wa paka waliochukua kutoka makao ya wanyama katika eneo lao kisha mama yao akawapigia simu kuwa paka wao wawili wa nyumba wamezaa watoto. Anasema kuwa Patchy, paka wa kwanza, ana watoto ambao idadi yao ni mara tatu ya watoto wa paka waliotwaa, huku Trixie, paka mwingine, ana watoto 12. Familia ya Doubtfire sasa ina watoto wangapi wa paka?
\textcolor{gray}{\textit{(Translation: The Doubtfire sisters are driving home with 7 kittens adopted from the local animal shelter when their mother calls to inform them that their two house cats have just had kittens. She says that Patchy, the first cat, has had thrice the number of adopted kittens, while Trixie, the other cat, has had 12. How many kittens does the Doubtfire family now have?)}}} \\
\midrule
\textbf{Reasoning trace} &
We need to parse the problem: It's in Swahili? Let's translate: ``Dada wa Doubtfire wanaendesha wakielekea nyumb.....So: ``Doubtfire's sister runs a home for stray cats (?) ...The phrase ``wakielekea nyumbani'' might be ``to bring home'' or ``to drive home''?... Actually ``wakielekea nyumbani'' might mean ``to bring home'' or ``to drive home''...But we need to decide which interpretation is correct.... But the question might ask for ``how many cat children does the Doubtfire family have?'' That includes the 7 kittens from the shelter and the new ones. So answer 55. But maybe the 7 cats are adult cats, not kittens. But the phrase ``watoto 7 wa paka'' is ambiguous. But I'd lean that it's 7 kittens.Thus answer: $\boxed{55}$. Let's provide final answer: 55. \\
\midrule
\textbf{Final response} &
Patchy has \(3 \times 12 = 36\) kittens, and Trixie has \(12\) kittens. The family already had \(7\) kitten-children from the shelter. 7 + 36 + 12 = 55.
$\boxed{55}$\textcolor{red}{\ding{55}} \\
\bottomrule
\end{tabular}
\caption{An example from gpt-oss-20b on PolyMath-Low where explicit uncertainty about an understanding failure appears in the reasoning trace but is omitted from the final response.}
\label{tab:verbalized_confusion_case_example}
\end{table*}

\section{Explicit Uncertainty in Understanding Failures}
\label{app:verbalized-confusion-evaluation}

Reasoning language models (RLMs) often exhibit difficulty in understanding queries written in languages with which they are less familiar~\citep{kang2026multilingualreasoninggapsemerge}. A representative example is shown in Table~\ref{tab:verbalized_confusion_case_example}. In this example, the model's reasoning trace explicitly expresses uncertainty about the Swahili wording and considers alternative interpretations, indicating difficulty in understanding the query. Despite recognizing this uncertainty, the model misinterprets a key condition in the problem: instead of using the fact that Patchy has three times the number of adopted kittens, i.e., $3 \times 7$, the model incorrectly computes the number of Patchy's kittens as $3 \times 12$. As a result, the model produces an incorrect answer due to an understanding failure.

However, this uncertainty is not reflected in the final response. Although the reasoning trace reveals that the model may not have properly understood the question, the final response presents the solution confidently without acknowledging this uncertainty. This is problematic because users typically observe only the final response, and therefore may not be aware that the answer is based on an uncertain or potentially incorrect understanding of the query. To better understand how often models explicitly express such uncertainty during response generation, we measure the \textit{explicit uncertainty rate} separately in the reasoning trace and in the final response. We measure explicit uncertainty in understanding-failure cases through a two-stage evaluator pipeline.

\paragraph{Stage 1.}
The first stage identifies examples in which the model's final response reflects a critical misunderstanding of the original question.
We conduct this analysis on PolyMath-Low~\citep{wang2025polymath}, using the same languages as in the main experiments.
For each of Qwen3-4B, Qwen3-8B~\cite{yang2025qwen3technicalreport}, gpt-oss-20b, and gpt-oss-120b~\cite{agarwal2025gpt}, we inspect only examples whose final answer is incorrect. Given the original question and the final response generated after the reasoning trace, we use gpt-5.5\footnote{\url{https://developers.openai.com/api/docs/models/gpt-5.5}} (\texttt{reasoning\_effort=high}) to determine whether the final response critically misunderstands the conditions specified in the question. Table~\ref{tab:verbalized_confusion_sample_counts} summarizes the results of Stage 1.
The evaluator prompt for Stage 1 is shown below.

\begin{tcolorbox}[
  colback=gray!5,
  colframe=gray!80!black,
  colbacktitle=gray!80!black,
  coltitle=white,
  title=Understanding-Failure Detection Prompt,
  breakable,
  enhanced
]
You are an evaluator of multilingual math question understanding.

Input: \\
- question: \{question\} \\
- response: \{response\} \\

Your task is to determine whether the response shows a critical misunderstanding of the question.

Return "fail": true only if the response shows a critical misunderstanding of the question that affects the solution.

Critical misunderstandings include: \\
- solving a different problem \\
- misunderstanding key quantities, objects, units, conditions, variables, or comparison targets \\
- ignoring or mistranslating multilingual text in a way that changes the meaning \\

Return "fail": false if the response understood the intended question well enough, even if it has minor arithmetic mistakes, wording issues, or an incomplete explanation.

Output format: \\
Return only a JSON object with a single boolean field: \\

\{"fail": true\} or \{"fail": false\}
\end{tcolorbox}

\begin{table}[!t]
\centering
\scriptsize
\setlength{\tabcolsep}{4pt}
\begin{tabular}{lccc}
\toprule
Model & Incorrect & Understanding failure & Failure rate (\%) \\
\midrule
Qwen3-4B & 433 & 358 & 82.7 \\
Qwen3-8B & 346 & 292 & 84.4 \\
gpt-oss-20b & 157 & 92 & 58.6 \\
gpt-oss-120b & 142 & 59 & 41.5 \\
\bottomrule
\end{tabular}
\caption{Stage 1 statistics for evaluating explicit uncertainty in understanding-failure cases on PolyMath-Low (1,250 examples).}
\label{tab:verbalized_confusion_sample_counts}
\end{table}

\paragraph{Stage 2.}
Among the examples identified in Stage 1, the second stage computes the explicit uncertainty rate. We define \emph{explicit uncertainty} as an explicit statement by the model that the wording, language, translation, or meaning of the question is confusing, ambiguous, difficult to parse, or uncertain.
This definition excludes ordinary step-by-step setup phrases such as ``let me understand the problem,'' wrong restatements, arithmetic errors, and incorrect answers, unless the model explicitly expresses uncertainty about the meaning of the question.
To examine where this phenomenon appears during response generation, we apply this check separately to the reasoning trace and to the final response.
Because this stage requires a direct surface-level judgment about whether such an explicit statement appears, we use gpt-oss-120b as the evaluator.
The evaluator prompts used in Stage 2 for reasoning traces and final responses are shown below.

\begin{tcolorbox}[
  colback=gray!5,
  colframe=gray!80!black,
  colbacktitle=gray!80!black,
  coltitle=white,
  title=Reasoning Trace Verbalized Confusion Detection Prompt,
  breakable,
  enhanced
]
You are a careful evaluator of model behavior. 
\\
You are given: \\
1. The original question. 2. The model's reasoning trace while answering that question.\\

Decide whether the model text explicitly verbalizes difficulty understanding the question.

Return true only when the model itself says the wording, language, translation, or meaning of the question is confusing, ambiguous, hard to parse, or uncertain. Examples include statements like "the wording is confusing", "I need to parse this more carefully", "does this mean X or Y?", or "I am not sure what the question is asking".

Return false for normal step-by-step reasoning, arithmetic mistakes, wrong answers, wrong restatements, or phrases like "let me understand/break down the problem" unless the model explicitly expresses confusion or uncertainty about the question's meaning.

Return exactly this JSON object:
\{
  "explicit\_understanding\_failure": true or false
\}

Original question:\\
<question>\\
\{question\}\\
</question>\\

Reasoning trace to inspect:\\
<reasoning\_trace>\\
\{model\_text\}\\
</reasoning\_trace>\\
\end{tcolorbox}

\begin{tcolorbox}[
  colback=gray!5,
  colframe=gray!80!black,
  colbacktitle=gray!80!black,
  coltitle=white,
  title=Final Response Verbalized Confusion Detection Prompt,
  breakable,
  enhanced
]
You are a careful evaluator of model behavior.\\

You are given:\\
1. The original question.
2. The model's final prediction for that question.

Decide whether the model text explicitly verbalizes difficulty understanding the question.

Return true only when the model itself says the wording, language, translation, or meaning of the question is confusing, ambiguous, hard to parse, or uncertain. Examples include statements like "the wording is confusing", "I need to parse this more carefully", "does this mean X or Y?", or "I am not sure what the question is asking".

Return false for normal step-by-step reasoning, arithmetic mistakes, wrong answers, wrong restatements, or phrases like "let me understand/break down the problem" unless the model explicitly expresses confusion or uncertainty about the question's meaning.

Return only a valid JSON object. Do not include Markdown, code fences, or extra text.
Return exactly this JSON object:
\{
  "explicit\_understanding\_failure": true or false
\}

Original question:\\
<question>\\
\{question\}\\
</question>\\

Final prediction to inspect:\\
<prediction>\\
\{model\_text\}\\
</prediction>\\
\end{tcolorbox}

\paragraph{Computing the explicit uncertainty rate.}
Let $\mathcal{F}_m$ be the set of examples for model $m$ whose final answer is incorrect and whose final response is judged to reflect an understanding failure by the first-stage evaluator. Let $u^{\mathrm{trace}}_{m,i}$ and $u^{\mathrm{final}}_{m,i}$ indicate whether the second-stage evaluator finds explicit uncertainty in the reasoning trace and final response, respectively.
We compute the explicit uncertainty rates as
\[
   \mathrm{Rate}^{\mathrm{trace}}_m
   =
   \frac{1}{|\mathcal{F}_m|}
   \sum_{i\in \mathcal{F}_m} u^{\mathrm{trace}}_{m,i},
\]
and
\[
   \mathrm{Rate}^{\mathrm{final}}_m
   =
   \frac{1}{|\mathcal{F}_m|}
   \sum_{i\in \mathcal{F}_m} u^{\mathrm{final}}_{m,i}.
\]
These quantities correspond to the explicit uncertainty rates in the reasoning trace and final response, respectively, as reported in Figure~\ref{fig:verbalized_failure_rates}.

%% file: latex/appendix/appendix_section_B.tex
\section{Dataset and Training Data Construction}
\label{app:dataset-training-data}

\paragraph{Evaluation datasets.}
We evaluate on PolyMath~\cite{wang2025polymath} and MMLU-ProX-Lite~\cite{xuanmmluprox} to cover both in-domain mathematical reasoning and out-of-domain STEM reasoning. PolyMath is a multilingual mathematical reasoning benchmark that covers multiple difficulty levels. We use the Low, Medium, and High splits: Low consists of K--12-style mathematical word problems, Medium contains exam-style and entry-level competition problems, and High focuses on more difficult competition problems. Each split provides parallel questions across languages, allowing direct comparison of multilingual reasoning performance. For PolyMath, we additionally include Yoruba and Zulu by translating the English questions with GPT-4.1, using the dataset translation prompt described below. MMLU-ProX-Lite is a compact version of MMLU-ProX with parallel multiple-choice questions across languages, including human-reviewed Yoruba and Zulu translations. Thus, evaluation on these unseen languages includes both machine-translated PolyMath questions and human-reviewed MMLU-ProX-Lite questions. We use its STEM-related subjects (\texttt{math, physics, chemistry, computer science,} and \texttt{engineering}) as an out-of-domain reasoning evaluation beyond the mathematical training data. For both benchmarks, we follow the evaluation prompt format of~\citet{kang2026multilingualreasoninggapsemerge}. 

\paragraph{Training source data.}
For training, we use the train split of \href{https://huggingface.co/datasets/agentica-org/DeepScaleR-Preview-Dataset}{agentica-org/DeepScaleR-Preview-Dataset}. Since DeepScaleR~\cite{deepscaler2025} is a mathematical reasoning dataset and PolyMath is also used for evaluation, we apply n-gram contamination filtering before sampling training examples. We normalize every English question in all PolyMath difficulty splits (Low, Medium, High, and Top) and build a character 13-gram index. We then compute the character n-gram Jaccard similarity between each DeepScaleR problem and each PolyMath question. A DeepScaleR problem is removed as contaminated if it has Jaccard similarity of at least $0.8$ with a PolyMath question and at least $10$ overlapping character 13-grams.

\paragraph{Training splits and languages.}
After filtering, we sample disjoint 10K-example sets for Stage I SFT and Stage II RL. The training languages are Arabic (ar), Thai (th), and Swahili (sw), and each split is balanced across these languages with an approximately $1{:}1{:}1$ ratio. We reserve 200 validation examples for model selection, also sampled in a language-balanced way, and use the remaining examples for training.

\paragraph{Translation for training data.}
We use GPT-4.1~\cite{openai2025gpt4.1} with decoding temperature $0$ to translate English DeepScaleR questions into the training languages. To filter obvious translation failures, we tokenize both the translated question and the English source question with the Qwen3~\cite{yang2025qwen3technicalreport} tokenizer and discard a translated question if its tokenized length is more than $15\times$ the tokenized length of the English source. 

\begin{tcolorbox}[
  colback=gray!5,
  colframe=gray!80!black,
  colbacktitle=gray!80!black,
  coltitle=white,
  title=Dataset Translation Prompt,
  breakable,
  enhanced
]
Translate the following mathematical question enclosed within <problem> and </problem> into \{language\_name\}.  
The text may contain mathematical notation and LaTeX formatting. You must strictly preserve:  \\
- All LaTeX math and commands EXACTLY as written, including inline math \texttt{\$...\$}, display math \texttt{\$\$...\$\$}, \texttt{\textbackslash(...\textbackslash)}, \texttt{\textbackslash[...\textbackslash]}, and any \texttt{\textbackslash begin\{...\}...\textbackslash end\{...\}} environments. \\
- All mathematical symbols, variables, numbers, operators, and equation labels.  \\

Provide only the translated problem without any additional explanation.
Wrap the translated output with <translated> and </translated> tags.
<problem>\{problem\}</problem>
\end{tcolorbox}

%% file: latex/appendix/appendix_section_C.tex
\section{Training Details}
\label{app:training-details}

\paragraph{Stage I SFT.}
Stage I uses AdamW~\cite{loshchilov2019decoupledweightdecayregularization} with learning rate $1\times10^{-5}$ and batch size $16$. The maximum prompt length is 2,560 tokens and the maximum response length is 12,288 tokens. We train for 3 epochs, evaluate every 200 steps, and select the checkpoint with the best validation accuracy-and-format reward, which is 1 only when the answer is accurate and the required output format is correct.

\paragraph{Stage II RL.}
Stage II uses group size $G=8$, AdamW with learning rate $2\times10^{-6}$, batch size $32$, mini-batch size $16$, rollout temperature $1.0$, and top-$p=1.0$. The maximum prompt length is 2,560 tokens and the maximum response length is 12,288 tokens. We use the asymmetric clipping parameters $\epsilon_l=0.2$ and $\epsilon_h=0.3$. We train for 150 steps, evaluate every 10 steps, and select the checkpoint with the best validation accuracy-and-format reward. We use this criterion for all training-based methods because it applies uniformly to methods with and without translator decisions, whereas mode-selection macro-F1 is defined only for methods that make a translator-call decision.

\paragraph{Training cost.}
Training \textsc{Luar} on two NVIDIA H200 GPUs takes approximately 5 hours for Stage I SFT and 13 hours for Stage II RL on Qwen3-4B, and 8 hours for Stage I SFT and 19 hours for Stage II RL on Qwen3-8B.

\paragraph{Format correctness.}
During training, the correctness indicator $c(y)$ requires both answer correctness and format correctness. A response is format-correct only if it contains exactly one \texttt{<think>} tag and one \texttt{</think>} tag, \texttt{<think>} appears before \texttt{</think>}, and the output starts with \texttt{<think>}. The final solution block after \texttt{</think>} must contain an extractable \texttt{\textbackslash boxed\{...\}} answer. If the response uses \texttt{<translator\_call>...</translator\_call>}, the translator call must be well formed, must not be nested, and its stripped content must contain at least five characters. The tags \texttt{<translator\_call>}, \texttt{</translator\_call>}, \texttt{<translator\_response>}, and \texttt{</translator\_response>} must not appear in the final solution block. For English queries, any translator tag makes the response format-invalid.

\begin{table}[t]
\centering
\scriptsize
\setlength{\tabcolsep}{5pt}
\begin{tabular}{ll}
\toprule
Setting & Value \\
\midrule
Optimizer & AdamW \\
Learning rate & $1\times10^{-5}$ \\
Batch size & 16 \\
Max prompt length & 2,560 \\
Max response length & 12,288 \\
Epochs & 3 \\
\bottomrule
\end{tabular}
\caption{SFT hyperparameters for \textsc{Luar} Stage I, \textsc{Boundary-SFT}, Q-Align, and TAPO Stage I. Sequence lengths are measured in tokens.}
\label{tab:sft_training_hyperparameters}
\end{table}

\begin{table}[t]
\centering
\scriptsize
\setlength{\tabcolsep}{5pt}
\begin{tabular}{ll}
\toprule
Setting & Value \\
\midrule
Optimizer & AdamW \\
Learning rate & $2\times10^{-6}$ \\
Batch size & 32 \\
PPO mini-batch size & 16 \\
Group size & 8 \\
Max prompt length & 2,560 \\
Max response length & 12,288 \\
Rollout temperature & 1.0 \\
Rollout top-$p$ & 1.0 \\
Clipping parameters & $\epsilon_l=0.2,\epsilon_h=0.3$ \\
Training steps & 150 \\
\bottomrule
\end{tabular}
\caption{RL hyperparameters for \textsc{Luar} Stage II, TAPO Stage II, and standard \textsc{GRPO}. Sequence lengths are measured in tokens.}
\label{tab:rl_training_hyperparameters}
\end{table}

\begin{table}[t]
\centering
\scriptsize
\setlength{\tabcolsep}{5pt}
\begin{tabular}{ll}
\toprule
Setting & Value \\
\midrule
Method & \textsc{ST} \\
Backbone & Frozen RLM \\
Features & Last-token final-layer hidden state \\
Classifier hidden size & $\{0,32,128,512,d/2,d\}$ \\
Learning rate & $\{10^{-3},10^{-4},10^{-5}\}$ \\
Optimizer & AdamW \\
Weight decay & 0.01 \\
Batch size & 64 \\
Epochs & 10 \\
Loss & Class-weighted cross-entropy \\
Model selection & Best validation macro-F1 \\
\bottomrule
\end{tabular}
\caption{Training hyperparameters for the \textsc{ST} prober baseline. Here, $d$ is the hidden-state dimension and hidden size 0 denotes a linear probe.}
\label{tab:prober_training_hyperparameters}
\end{table}

%% file: latex/appendix/appendix_section_D.tex
\section{Evaluation Details}
\label{app:evaluation-details}
\subsection{Evaluation Protocol}

\paragraph{Decoding parameters.} We average metrics over three random seeds, using temperature 0.6, top-$p=0.95$, top-$k=50$, a maximum prompt length of 2,560 tokens, and a maximum response length of 12,288 tokens.

\paragraph{Mode-selection labels.}
We construct evaluation labels using the procedure in \S\ref{sec:boundary-estimation} separately for each method's evaluated checkpoint. For trained methods, labels are generated after training using the selected fine-tuned checkpoint; methods without backbone training use the base model. These labels are used only for evaluation and remain fixed throughout evaluation. Thus, mode-selection macro-F1 measures routing decisions relative to each method's own estimated translation benefit.

\paragraph{Translation at evaluation time.}
When evaluation requires translation, we use GPT-4.1~\cite{openai2025gpt4.1} with the tool-calling translation prompt shown below. This applies to both external translation baselines and translator calls made by \textsc{Luar}. We implement translator-call execution with VerlTool~\cite{jiang2025verltool}: when the model emits a well-formed \texttt{translator\_call} span, the environment appends the GPT-4.1 translation as a \texttt{translator\_response} and continues decoding from the updated context. In contrast to training-time translator-augmented rollouts, evaluation uses GPT-based translations rather than reference English questions.

\begin{tcolorbox}[
  colback=gray!5,
  colframe=gray!80!black,
  colbacktitle=gray!80!black,
  coltitle=white,
  title=Tool-Calling Translation Prompt,
  breakable,
  enhanced
]
Translate the following text enclosed within \texttt{<instruction>} and \texttt{</instruction>} into English. \\
The text may contain mathematical notation and LaTeX formatting. In this case, you must strictly preserve: \\
- All LaTeX math and commands EXACTLY as written. \\
- All mathematical symbols, variables, numbers, operators, and equation labels. \\

Provide only the translated text without any additional explanation. \\
Wrap the translated text with \texttt{<translated>} and \texttt{</translated>} tags. \\

\texttt{<instruction>\{input\_text\}
</instruction>}
\end{tcolorbox}

\paragraph{Answer extraction and verification.}
We follow~\citet{kang2026multilingualreasoninggapsemerge} for answer extraction and verification. For PolyMath~\cite{wang2025polymath}, we extract answers from the \texttt{\textbackslash boxed\{\}} pattern and verify them against the gold answers using Math-Verify.\footnote{\url{https://github.com/huggingface/Math-Verify}} For MMLU-ProX-Lite~\cite{xuanmmluprox}, we use a regular-expression matcher to extract the final multiple-choice answer and compare it with the gold option.

\input{latex/appendix/appendix_inference_efficiency}

%% file: latex/appendix/appendix_inference_efficiency.tex
\subsection{Inference Efficiency Measurement}
\label{app:inference_efficiency_measurement}

We report inference efficiency for Qwen3-4B using a single evaluation run over 6,320 queries: 1,250 queries from each PolyMath difficulty level and 2,570 from MMLU-ProX-Lite. Overall token means and cost estimates are weighted by query count rather than averaging dataset means equally. Table~\ref{tab:appendix_inference_efficiency_details} reports the detailed token counts and efficiency estimates.

\begin{table*}[t]
\centering
\small
\setlength{\tabcolsep}{5pt}
\caption{Detailed inference efficiency from the single Qwen3-4B evaluation run. Token counts are means over all queries, and GPT-4.1 inputs include translation instructions. Costs are in USD per query. Rounding can prevent cost components from summing exactly.}
\label{tab:appendix_inference_efficiency_details}
\textbf{(a) Mean token counts per query}\\[3pt]
\begin{tabular}{lrrrrr}
\toprule
\textbf{Method} & \multicolumn{2}{c}{\textbf{Qwen3-4B tokens}} & \multicolumn{2}{c}{\textbf{GPT-4.1 tokens}} & \textbf{Translator use (\%)} \\
\cmidrule(lr){2-3}\cmidrule(lr){4-5}
& \textbf{Input} & \textbf{Output} & \textbf{Input} & \textbf{Output} & \\
\midrule
Base & 289 & 9,251 & 0 & 0 & 0.0 \\
\textsc{ST} & 606 & 10,730 & 109 & 65 & 30.0 \\
\textsc{Full-Translation} & 819 & 9,203 & 290 & 158 & 100.0 \\
\rowcolor{gray!12}\textbf{\textsc{Luar}} & 465 & 6,623 & 60 & 30 & 20.5 \\
\midrule
\textsc{Luar} (Base output length) & 465 & 9,251 & 60 & 30 & 20.5 \\
\bottomrule
\end{tabular}
\\[8pt]
\textbf{(b) Latency and estimated costs in USD per query}\\[3pt]
\begin{tabular}{lrrrr}
\toprule
\textbf{Method} & \textbf{Latency (s)} & \textbf{Total (relative to Base)} & \textbf{Reasoning} & \textbf{Translation} \\
\midrule
Base & 11.77 & \$0.00191 (1.00$\times$) & \$0.00191 & \$0.00000 \\
\textsc{ST} & 14.09 & \$0.00300 (1.57$\times$) & \$0.00227 & \$0.00073 \\
\textsc{Full-Translation} & 11.97 & \$0.00385 (2.02$\times$) & \$0.00200 & \$0.00184 \\
\rowcolor{gray!12}\textbf{\textsc{Luar}} & 9.32 & \$0.00178 (0.93$\times$) & \$0.00142 & \$0.00036 \\
\midrule
\textsc{Luar} (Base output length) & -- & \$0.00230 (1.21$\times$) & \$0.00194 & \$0.00036 \\
\bottomrule
\end{tabular}
\end{table*}

\paragraph{API cost estimation.}
We estimate reasoning cost from Qwen3-4B input and output token counts at \$0.20 per million tokens for both input and output (Fireworks), and translation cost from GPT-4.1 input and output tokens at \$2 and \$8 per million tokens, respectively. All costs are in USD per query.

Token counts are reconstructed from evaluation traces and prompts using the Qwen3-4B tokenizer and the GPT-4.1 tiktoken encoding. Qwen3-4B input includes the chat-formatted prompt, supplied translator-prefill context where applicable, and repeated inference stages. GPT-4.1 input includes the translation instruction wrapper and source text, and its output is the extracted translated response. The costs are
\begin{align}
C_{\mathrm{reason}}(q) &= \frac{0.20\,(n^{Q}_{\mathrm{in}}(q)+n^{Q}_{\mathrm{out}}(q))}{10^6},\\
C_{\mathrm{trans}}(q) &= \frac{2\,n^{T}_{\mathrm{in}}(q)+8\,n^{T}_{\mathrm{out}}(q)}{10^6},\\
C_{\mathrm{total}}(q) &= C_{\mathrm{reason}}(q)+C_{\mathrm{trans}}(q),
\end{align}
where, for query $q$, $n^{Q}_{\mathrm{in}}(q)$ and $n^{Q}_{\mathrm{out}}(q)$ are the Qwen3-4B input and output token counts, and $n^{T}_{\mathrm{in}}(q)$ and $n^{T}_{\mathrm{out}}(q)$ are the corresponding GPT-4.1 counts. Costs are averaged over all queries in the run. Translator usage is the fraction of queries with at least one translation request.

\paragraph{Latency measurement.}
Reasoning execution is measured on one NVIDIA A100 GPU with 80 GB of memory using vLLM. Qwen3-4B model loading, warm-up, and \textsc{ST} prober execution are included. We sum the elapsed times for Qwen3-4B reasoning and GPT-4.1 translation, then divide by the total number of evaluation queries to compute latency per query:
\begin{equation}
L = \frac{T_{\mathrm{Qwen\ batch}} + T_{\mathrm{translation\ batch}}}{N_{\mathrm{all\ queries}}}.
\end{equation}
GPT-4.1 translation requests are processed in batches with a maximum of eight concurrent API requests.

\paragraph{Translator usage and translation cost.}
To interpret translation costs, we consider how translator calls are distributed across datasets. These distributions differ across methods, so translator usage and translation cost need not be proportional. For example, \textsc{ST} concentrates its calls on MMLU-ProX-Lite, which has longer translation inputs and outputs than PolyMath.

\paragraph{Reference cost estimate.}
For reference, Table~\ref{tab:appendix_inference_efficiency_details} also includes \textsc{Luar} (Base output length). This hypothetical cost estimate replaces \textsc{Luar}'s mean reasoning output length with Base's while keeping all other token counts unchanged. Under this setting, \textsc{Luar}'s estimated total cost remains lower than those of \textsc{ST} and \textsc{Full-Translation}, indicating that its cost advantage persists when its output length is matched to Base.

%% file: latex/appendix/appendix_section_E.tex
\section{Baseline Implementation Details}
\label{app:baseline-details}

\paragraph{\textsc{ST}.}
\label{app:st-implementation}
We construct prober labels following the translator-usefulness estimation procedure in \S~\ref{sec:boundary-estimation}. A prober is a lightweight classifier trained on frozen hidden representations of the backbone model to predict whether translation is useful for a query. 
Specifically, for each DeepScaleR training question, we construct the translator usefulness label following the procedure in \S~\ref{sec:boundary-estimation}. Following~\citet{kang2026multilingualreasoninggapsemerge}, we train the prober with class-weighted cross-entropy loss and use their grid-search protocol over learning rates $\{10^{-3}, 10^{-4}, 10^{-5}\}$ and hidden-layer sizes $\{0, 32, 128, 512, d/2, d\}$, where $d$ is the hidden-state dimensionality and $0$ corresponds to a linear probe. \textsc{ST} extracts the final-layer hidden state at the last token of the reasoning trace. The backbone language model is frozen and only the prober is trained. The \textsc{ST} pipeline completes direct reasoning before applying the prober and reruns reasoning on the translated question if the prober predicts translation usefulness. The prober search space and training hyperparameters are summarized in Table~\ref{tab:prober_training_hyperparameters}.

\paragraph{\textsc{Boundary-SFT}.}
\textsc{Boundary-SFT} uses the same rollout-based translation-usefulness labels as \textsc{Luar}, constructed as described in \S~\ref{sec:boundary-estimation}. For each question, we select a translator-augmented target only when the translator-augmented accuracy is at least $0.5$ higher than the direct accuracy. Otherwise, we select a direct target. We retain only sampled responses that give the correct answer and conform to the required output format. The SFT hyperparameters are given in Table~\ref{tab:sft_training_hyperparameters}.

\paragraph{\textsc{Q-Align}.}
\textsc{Q-Align}~\cite{zhu-etal-2024-question} improves multilingual reasoning by training the model to translate non-English reasoning questions into English, thereby aligning non-English questions with their English counterparts and enabling transfer from English reasoning ability. Since our backbone is already a post-trained RLM, we avoid directly optimizing a separate question-translation objective, which may disrupt its learned reasoning behavior by encouraging translation-only outputs. Instead, we adapt \textsc{Q-Align} by inserting the gold English translation into the reasoning trace and training the model to solve the problem conditioned on it. We construct targets from the same sampled DeepScaleR training questions used for \textsc{Luar}. During sampling, we force the model to condition on the English translation of the problem before solving, using the prefix below. We retain only sampled responses that are answer-correct and format-valid. The SFT hyperparameters are summarized in Table~\ref{tab:sft_training_hyperparameters}.

\begin{tcolorbox}[
  colback=gray!5,
  colframe=gray!80!black,
  colbacktitle=gray!80!black,
  coltitle=white,
  title=Q-Align Forced Prefix,
  breakable,
  enhanced
]
<think>
Okay, let me first translate the question into English to make sure I
understand it correctly.

English translation:\\
\{prompt\_text\_in\_english\}

Now that the question is clear in English, I will solve the problem step by step and derive the final answer.

\end{tcolorbox}

\paragraph{\textsc{TAPO}.}
\textsc{TAPO}~\cite{huang2026tapo} was originally proposed for instruction-tuned models, where the model is trained to follow an explicit understand-then-reason format: first translate the problem into English, and then solve it based on the translation. Since our backbone is already a post-trained RLM, directly imposing the original pipeline may disrupt its learned reasoning format and behavior. We therefore adapt \textsc{TAPO} by preserving its explicit translation-augmented reasoning structure, while introducing it into the RLM through a two-stage training pipeline.

We use the original \textsc{TAPO} system prompt for both Stage I sampling and Stage II RL.
Stage I transfers the \textsc{TAPO} reasoning format to the RLM: we sample DeepScaleR responses under the \textsc{TAPO} system prompt with a forced English-translation prefix, so that the resulting reasoning traces begin with an explicit English translation block before continuing to reasoning and the final solution. Only answer-correct and format-valid sampled responses are retained for SFT.
Stage II then optimizes this behavior using the \textsc{TAPO} reward and step-level advantage formulation, which decouples credit assignment for translation and reasoning tokens.
We use the same RL hyperparameters as \textsc{Luar}, except for the reward and advantage formulation.
We use ChrF++~\cite{popovic-2017-chrf} score\footnote{
We compute ChrF++ using the Hugging Face \texttt{evaluate} implementation:
\url{https://github.com/huggingface/evaluate/blob/main/metrics/chrf/README.md},
with \texttt{word\_order=2} and \texttt{lowercase=True}.
} as the translation reward and set the advantage interpolation coefficient $\alpha$ to $0.25$, following the original \textsc{TAPO} setup. The SFT and RL hyperparameters are listed in Tables~\ref{tab:sft_training_hyperparameters} and~\ref{tab:rl_training_hyperparameters}.

\begin{tcolorbox}[
  colback=gray!5,
  colframe=gray!80!black,
  colbacktitle=gray!80!black,
  coltitle=white,
  title=TAPO Forced Translation Prefix,
  breakable,
  enhanced
]
<think>\\
<english\_translation>\\
\{prompt\_text\_in\_english\}\\
</english\_translation>
\end{tcolorbox}

\paragraph{Standard \textsc{GRPO}.}
The standard \textsc{GRPO}~\cite{shao2024deepseekmathpushinglimitsmathematical} baseline is trained on the same RL data as \textsc{Luar} and uses the same RL hyperparameters, as shown in Table~\ref{tab:rl_training_hyperparameters}. The only difference is the reward: \textsc{GRPO} uses an accuracy-and-format reward ($c(y)$) without \textsc{Luar}'s translation-usefulness reward signal.

\section{Prompt Formats}
\label{app:prompt-formats}

This appendix lists the prompt and tool formats used for translator-augmented reasoning and prompt-based translator-call baselines.

\subsection{Translator-Augmented Reasoning}

\begin{tcolorbox}[
  colback=gray!5,
  colframe=gray!80!black,
  colbacktitle=gray!80!black,
  coltitle=white,
  title=Translator-Augmented Reasoning Prefix,
  breakable,
  enhanced
]
<think>\\
Okay, let's see. As I read the question, some parts are difficult for me to fully understand. I'll translate it into English to avoid any
misunderstanding.\\
Let me call the translator to get an English version of the question:\\
<translator\_call>\\
\{prompt\_text\_in\_original\_language\}\\
</translator\_call>\\
<translator\_response>\\
\{prompt\_text\_in\_english\}\\
</translator\_response>\\
Now that I have the translated version, I can proceed with the reasoning
step by step.
\end{tcolorbox}

\subsection{Self-Assessment Baseline}
\begin{tcolorbox}[
  colback=gray!5,
  colframe=gray!80!black,
  colbacktitle=gray!80!black,
  coltitle=white,
  title=Self-Assessment Baseline Prompt,
  breakable,
  enhanced
]
You are a careful and reliable reasoning assistant. Solve the given question step by step.

After solving the problem, assess whether you confidently understood the
question itself in its original language. This judgment should reflect
whether the question was understandable, not whether your answer is
correct.

If the question was written in an unfamiliar language, difficult to
interpret, or if you are uncertain whether you understood it correctly,
choose
understandable: no.

Otherwise choose understandable: yes.

At the very end of your output, format your response exactly as follows:

Answer: <your answer>
Understandable: yes/no
\end{tcolorbox}

\subsection{Native-Tool-Use Baseline}
\begin{tcolorbox}[
  colback=gray!5,
  colframe=gray!80!black,
  colbacktitle=gray!80!black,
  coltitle=white,
  title=Native-Tool-Use Baseline Tool Description,
  breakable,
  enhanced
]
Tool name: translate\_query\_to\_english

Description:
Translate the input query into English when necessary for accurate
understanding. You may call this tool if the query is in a language you are not confident in interpreting. Avoid using it when the query is already in English or when its meaning is clear.

Parameter:
input\_query: The input query to translate into English.
\end{tcolorbox}

%% file: latex/appendix/appendix_section_F.tex
\section{Additional Experimental Results}
\subsection{Robustness to Translation-Usefulness Thresholds}
\label{app:threshold_robustness}

\begin{figure}[t]
    \centering
    \includegraphics[width=\columnwidth]{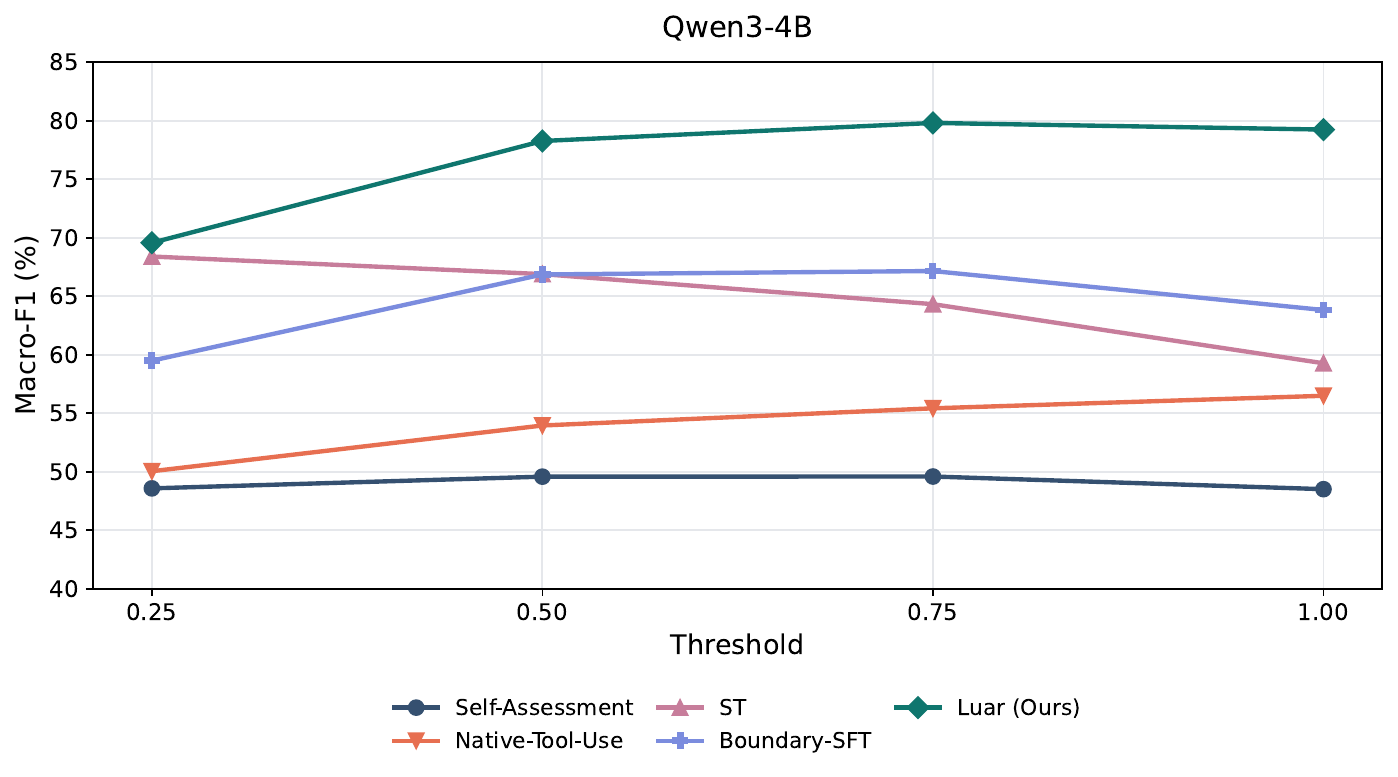}
    \caption{Mode-selection macro-F1 under different boundary thresholds on Qwen3-4B.}
    \label{fig:macro_f1_threshold_qwen3_4b}
\end{figure}

\begin{figure}[t]
    \centering
    \includegraphics[width=\columnwidth]{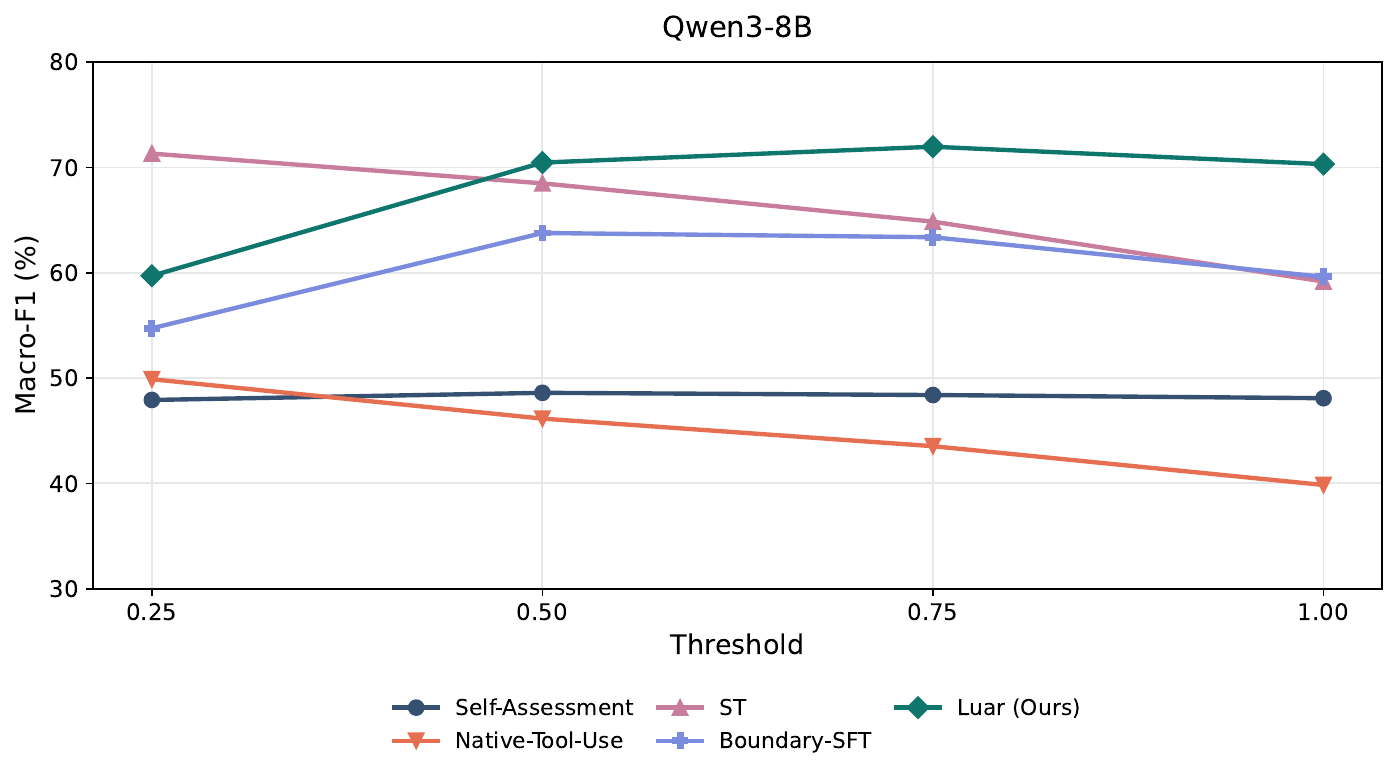}
    \caption{Mode-selection macro-F1 under different boundary thresholds on Qwen3-8B.}
    \label{fig:macro_f1_threshold_qwen3_8b}
\end{figure}
Our main experiments define the translation-usefulness label with
$\delta=0.5$ in Eq.~\ref{eq:boundary-label} for evaluation. To examine whether
\textsc{Luar}'s mode-selection advantage holds beyond this particular
evaluation definition, we additionally evaluate macro-F1 under alternative
boundary definitions. Specifically, we vary
$\delta \in \{0.25, 0.5, 0.75, 1.0\}$. At $\delta=1.0$, a query is
labeled translation-useful only when $t_{\mathrm{acc}}(q)=1.0$ and
$d_{\mathrm{acc}}(q)=0.0$, corresponding to all sampled translator-augmented
responses being correct and all sampled direct responses being incorrect.

Figures~\ref{fig:macro_f1_threshold_qwen3_4b} and~\ref{fig:macro_f1_threshold_qwen3_8b} show that \textsc{Luar}
maintains strong mode-selection performance across these alternative
boundary definitions. In most settings, \textsc{Luar} achieves the best or
highly competitive macro-F1 compared with prompt-based, external
detector-based, and SFT-based baselines. The advantage is especially clear
under stricter thresholds, where positive examples correspond to cases in
which translation yields a large empirical improvement over direct
reasoning. These results show that \textsc{Luar} remains effective not only
under the main $\delta=0.5$ definition, but also when translation usefulness
is evaluated with more relaxed or stricter criteria.

\subsection{Decision Diagnostics}
\label{app:decision-diagnostics}

\begin{figure*}[t]
\centering
\includegraphics[width=\textwidth]{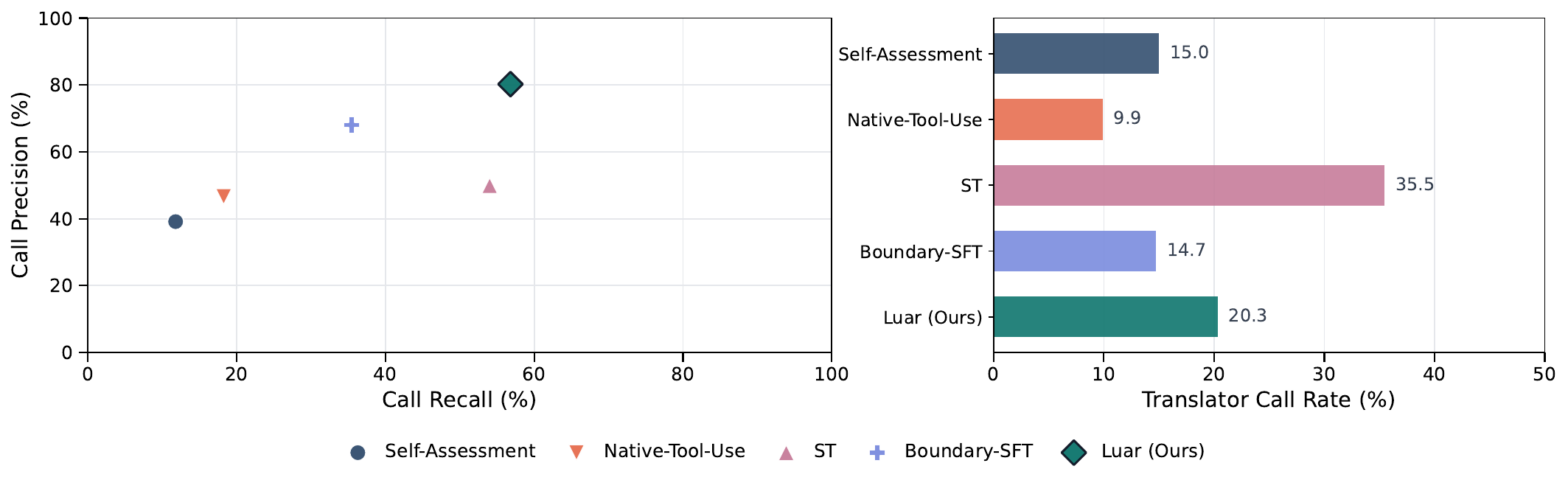}
\caption{Precision and recall for translator-call decisions on Qwen3-4B. \textsc{Luar} improves decision quality while keeping translator use selective across datasets.}
\label{fig:precision_recall_qwen3_4b}
\end{figure*}

\begin{figure*}[t]
\centering
\includegraphics[width=\textwidth]{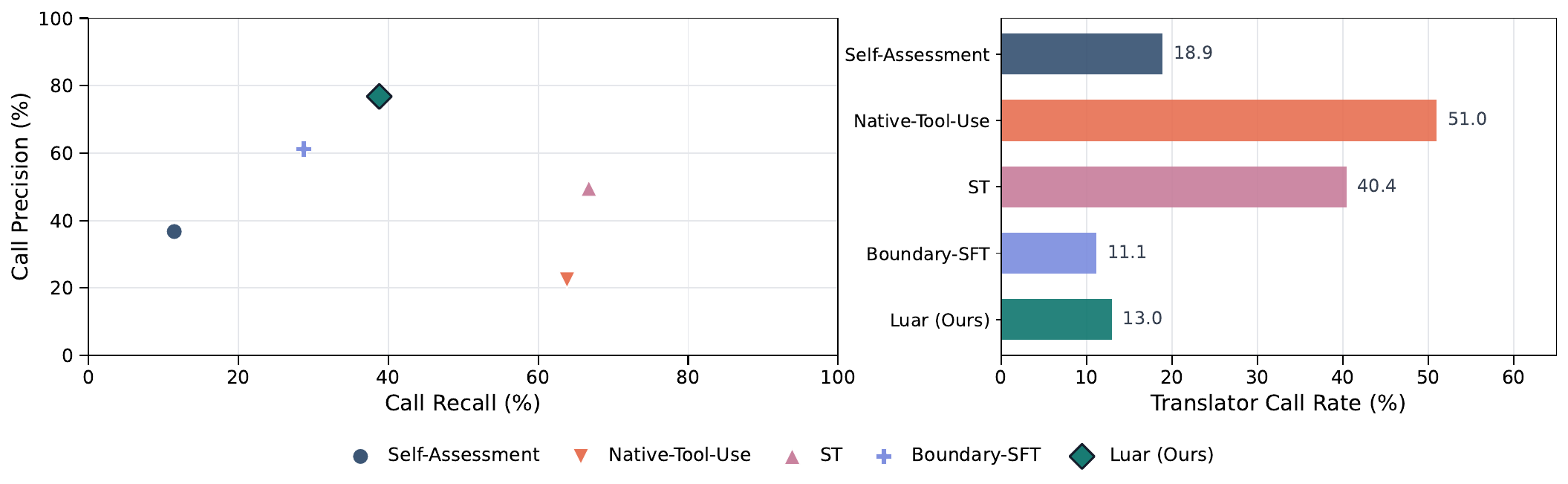}
\caption{Precision and recall for translator-call decisions on Qwen3-8B. \textsc{ST} attains high recall through substantially higher translator usage than \textsc{Luar}, but this comes with lower precision.}
\label{fig:precision_recall_qwen3_8b}
\end{figure*}

Figures~\ref{fig:precision_recall_qwen3_4b} and~\ref{fig:precision_recall_qwen3_8b} decompose translator-call decisions into precision, recall, and overall call rate.

\subsection{Per-Language Results}
\label{app:per-language-results}

Tables~\ref{tab:per_language_accuracy_base_relative_colored_polymath_low}--\ref{tab:per_language_accuracy_base_relative_colored_mmlu_prox_lite} report per-language task accuracy for Qwen3-4B and Qwen3-8B across all evaluation datasets.

\input{latex/table/appendix_per_language_accuracy}

\subsection{Within-Language Translator-Call Behavior}
\label{app:within-language-routing}

To investigate how \textsc{Luar} calls the translator within the same language, we analyze its call behavior in relation to translation usefulness. Following \S\ref{sec:boundary-estimation}, we estimate direct accuracy $d_{\mathrm{acc}}(q)$ and translator-augmented accuracy $t_{\mathrm{acc}}(q)$ from sampled base Qwen3-4B responses across all evaluation datasets and languages. We define two groups: \emph{direct-solvable} queries satisfy $d_{\mathrm{acc}}(q)=1$ and $t_{\mathrm{acc}}(q)\leq d_{\mathrm{acc}}(q)$, while \emph{translation-rescued} queries satisfy $d_{\mathrm{acc}}(q)=0$ and $t_{\mathrm{acc}}(q)=1$. These groups distinguish queries that the base model consistently solves directly from those it consistently solves only with translation. We compare \textsc{Luar}'s within-language call rates on these two groups, excluding all other queries.

\input{latex/table/appendix_within_language_call_rates}

Table~\ref{tab:within_language_call_rates} shows substantially higher translator-call rates for translation-rescued queries than for direct-solvable queries overall (72.1\% vs. 8.3\%). The same pattern appears in several languages. For example, the call rates for translation-rescued and direct-solvable queries are 61.3\% vs. 22.8\% in Yoruba and 94.3\% vs. 59.0\% in Zulu.

Although routing is not perfectly selective, these within-language differences suggest potential benefits over routing based solely on language identity. Figure~\ref{fig:translator_call_rate} shows that \textsc{Luar} frequently calls the translator for unseen Yoruba and Zulu, where the base model struggles to reason directly, without prior estimates of translation benefit for either language. A router based only on language identity would need such estimates to target those languages.

%% file: latex/table/appendix_per_language_accuracy.tex
\begin{table*}[t]
\centering
\small
\setlength{\tabcolsep}{4.5pt}
\begin{tabular}{lccccccccccc}
\toprule
\textbf{Method} 
& \multicolumn{3}{c}{\textbf{Seen}} 
& \multicolumn{7}{c}{\textbf{Unseen}} 
& \textbf{Avg.} \\
\cmidrule(lr){2-4}\cmidrule(lr){5-11}
& \textbf{ar} & \textbf{th} & \textbf{sw}
& \textbf{en} & \textbf{zh} & \textbf{ko} & \textbf{bn} & \textbf{te} & \textbf{yo} & \textbf{zu}
& \\
\midrule
\textsc{Qwen3-4B} & 89.33 & 83.73 & 32.53 & 95.73 & 86.93 & 89.60 & 82.93 & 66.67 & 8.53 & 8.00 & 64.40 \\
\midrule
\textsc{Full-Translation} & \cellcolor{green!3}91.20 & \cellcolor{green!12}90.40 & \cellcolor{green!35}84.80 & 95.73 & \cellcolor{green!10}92.80 & \cellcolor{green!2}90.93 & \cellcolor{green!14}91.20 & \cellcolor{green!27}81.87 & \cellcolor{green!35}78.40 & \cellcolor{green!35}88.53 & \cellcolor{green!35}88.59 \\
\specialrule{0.10em}{0.05em}{0.05em}
\textsc{Self-Assessment} & \cellcolor{green!2}90.67 & \cellcolor{red!2}82.40 & \cellcolor{red!3}30.67 & \cellcolor{red!1}94.93 & \cellcolor{green!2}88.00 & \cellcolor{red!1}89.33 & \cellcolor{red!1}82.67 & \cellcolor{green!1}66.93 & \cellcolor{red!6}5.07 & \cellcolor{red!6}4.53 & \cellcolor{red!2}63.52  \\
\textsc{Native-Tool-Use} & \cellcolor{green!1}89.87 & \cellcolor{red!1}83.47 & \cellcolor{green!14}40.53 & \cellcolor{red!3}94.13 & \cellcolor{green!3}88.80 & 89.60 & \cellcolor{green!1}83.73 & \cellcolor{green!6}70.13 & \cellcolor{green!35}29.33 & \cellcolor{green!35}33.07 & \cellcolor{green!10}70.27 \\
\textsc{ST} & \cellcolor{green!1}90.13 & \cellcolor{green!3}85.33 & \cellcolor{green!35}52.53 & \cellcolor{green!1}96.00 & \cellcolor{green!2}88.27 & \cellcolor{green!1}89.87 & \cellcolor{green!2}84.27 & \cellcolor{green!6}69.87 & \cellcolor{green!35}58.93 & \cellcolor{green!35}78.40 & \cellcolor{green!26}79.36 \\
\textsc{Boundary-SFT} & \cellcolor{green!1}90.00 & \cellcolor{red!1}83.20 & \cellcolor{green!35}54.80 & \cellcolor{red!2}94.80 & \cellcolor{red!6}83.60 & \cellcolor{red!9}84.40 & \cellcolor{red!6}79.60 & \cellcolor{green!5}69.60 & \cellcolor{green!35}30.80 & \cellcolor{green!35}71.20 & \cellcolor{green!17}74.20 \\
\midrule
\textsc{Q-Align} & \cellcolor{red!3}87.73 & \cellcolor{green!1}84.00 & \cellcolor{green!10}38.40 & 95.73 & 86.93 & \cellcolor{red!10}84.00 & \cellcolor{red!10}77.33 & \cellcolor{red!6}63.20 & \cellcolor{red!10}2.67 & \cellcolor{red!7}4.27 & \cellcolor{red!3}62.43 \\
\textsc{GRPO} & \cellcolor{red!2}88.00 & \cellcolor{red!1}83.20 & \cellcolor{red!5}29.60 & \cellcolor{red!1}94.93 & \cellcolor{green!1}87.73 & \cellcolor{red!1}89.33 & \cellcolor{red!4}80.80 & \cellcolor{red!2}65.60 & \cellcolor{green!5}11.20 & \cellcolor{red!2}6.67 & \cellcolor{red!1}63.71 \\
\textsc{TAPO} & \cellcolor{red!11}83.20 & \cellcolor{red!5}81.07 & \cellcolor{green!12}39.20 & \cellcolor{red!1}95.47 & \cellcolor{red!2}85.87 & \cellcolor{red!9}84.53 & \cellcolor{red!19}72.00 & \cellcolor{red!9}61.33 & \cellcolor{red!8}3.73 & \cellcolor{red!4}5.60 & \cellcolor{red!6}61.20 \\
\midrule
\rowcolor{gray!12} \textsc{\textbf{Luar}} & \cellcolor{green!1}89.60 & \cellcolor{green!2}85.07 & \cellcolor{green!35}89.60 & 95.73 & \cellcolor{green!4}89.07 & \cellcolor{red!5}86.67 & \cellcolor{red!2}81.60 & \cellcolor{green!2}68.00 & \cellcolor{green!35}56.27 & \cellcolor{green!35}87.20 & \cellcolor{green!32}\textbf{82.88} \\
\midrule
\textsc{Qwen3-8B} & 90.40 & 89.87 & 62.67 & 96.53 & 89.33 & 91.47 & 89.33 & 77.60 & 13.60 & 15.73 & 71.65 \\
\midrule
\textsc{Full-Translation} & \cellcolor{green!3}92.00 & \cellcolor{green!5}92.53 & \cellcolor{green!35}88.53 & 96.53 & \cellcolor{green!4}91.47 & \cellcolor{red!1}90.67 & \cellcolor{green!4}91.73 & \cellcolor{green!14}85.33 & \cellcolor{green!35}78.67 & \cellcolor{green!35}89.33 & \cellcolor{green!32}89.68 \\
\specialrule{0.10em}{0.05em}{0.05em}
\textsc{Self-Assessment} & \cellcolor{green!6}93.60 & \cellcolor{green!3}91.73 & \cellcolor{green!1}63.47 & 96.53 & \cellcolor{green!1}89.60 & \cellcolor{red!1}91.20 & \cellcolor{red!2}88.27 & \cellcolor{red!2}76.27 & \cellcolor{red!6}10.40 & \cellcolor{green!3}17.33 & \cellcolor{green!1}71.84 \\
\textsc{Native-Tool-Use} & \cellcolor{red!35}63.47 & \cellcolor{red!35}61.07 & \cellcolor{red!1}62.13 & \cellcolor{red!20}85.07 & \cellcolor{red!12}82.67 & \cellcolor{red!25}77.07 & \cellcolor{red!35}67.20 & \cellcolor{red!30}60.53 & \cellcolor{green!35}65.33 & \cellcolor{green!35}78.40 & \cellcolor{red!2}70.29 \\
\textsc{ST} & 90.40 & \cellcolor{green!1}90.67 & \cellcolor{green!28}78.67 & \cellcolor{red!1}96.27 & \cellcolor{green!1}89.87 & \cellcolor{red!2}90.40 & \cellcolor{red!1}88.53 & \cellcolor{green!7}81.33 & \cellcolor{green!35}63.47 & \cellcolor{green!35}69.07 & \cellcolor{green!21}\textbf{83.87} \\
\textsc{Boundary-SFT} & \cellcolor{red!1}89.60 & 89.87 & \cellcolor{green!2}63.73 & \cellcolor{green!1}96.80 & \cellcolor{red!2}88.27 & \cellcolor{red!3}89.60 & \cellcolor{red!6}86.13 & \cellcolor{red!2}76.27 & \cellcolor{green!35}34.13 & \cellcolor{green!35}50.40 & \cellcolor{green!8}76.48 \\
\midrule
\textsc{Q-Align} & \cellcolor{red!2}89.33 & \cellcolor{red!1}89.33 & \cellcolor{green!2}64.00 & \cellcolor{green!1}97.33 & \cellcolor{red!3}87.73 & \cellcolor{red!6}88.27 & \cellcolor{red!14}81.60 & \cellcolor{red!6}74.13 & \cellcolor{red!11}7.47 & \cellcolor{red!14}8.00 & \cellcolor{red!5}68.72 \\
\textsc{GRPO} & 90.40 & \cellcolor{green!1}90.67 & 62.67 & 96.53 & \cellcolor{green!1}89.60 & \cellcolor{red!1}91.20 & \cellcolor{red!5}86.40 & \cellcolor{red!4}75.20 & \cellcolor{red!11}7.47 & \cellcolor{red!3}13.87 & \cellcolor{red!2}70.40 \\
\textsc{TAPO} & \cellcolor{red!12}83.73 & \cellcolor{red!13}82.67 & \cellcolor{red!35}38.93 & \cellcolor{red!2}95.47 & \cellcolor{red!8}84.80 & \cellcolor{red!12}84.80 & \cellcolor{red!31}71.73 & \cellcolor{red!29}60.80 & \cellcolor{red!15}4.80 & \cellcolor{red!21}3.47 & \cellcolor{red!18}61.12 \\
\midrule
\rowcolor{gray!12} \textsc{\textbf{Luar}} & \cellcolor{green!1}90.67 & \cellcolor{green!1}90.13 & \cellcolor{green!24}76.27 & \cellcolor{green!1}96.80 & \cellcolor{red!4}87.20 & \cellcolor{red!4}89.33 & \cellcolor{red!5}86.40 & \cellcolor{red!1}76.80 & \cellcolor{green!35}40.27 & \cellcolor{green!35}69.87 & \cellcolor{green!15}80.37 \\
\bottomrule
\end{tabular}
\caption{Per-language task accuracy (\%) on PolyMath-Low. Numbers are mean accuracies over seeds. Bold indicates the best average among methods that do not translate every input. Cell colors indicate changes from the corresponding base-model row, clipped at $\pm$20 points.}
\label{tab:per_language_accuracy_base_relative_colored_polymath_low}
\end{table*}

\begin{table*}[t]
\centering
\small
\setlength{\tabcolsep}{4.5pt}
\begin{tabular}{lccccccccccc}
\toprule
\textbf{Method} 
& \multicolumn{3}{c}{\textbf{Seen}} 
& \multicolumn{7}{c}{\textbf{Unseen}} 
& \textbf{Avg.} \\
\cmidrule(lr){2-4}\cmidrule(lr){5-11}
& \textbf{ar} & \textbf{th} & \textbf{sw}
& \textbf{en} & \textbf{zh} & \textbf{ko} & \textbf{bn} & \textbf{te} & \textbf{yo} & \textbf{zu}
& \\
\midrule
\textsc{Qwen3-4B} & 37.87 & 35.20 & 25.33 & 37.07 & 36.80 & 36.80 & 35.20 & 35.20 & 20.80 & 20.27 & 32.05 \\
\midrule
\textsc{Full-Translation} & \cellcolor{red!2}36.80 & \cellcolor{green!3}36.80 & \cellcolor{green!18}35.47 & 37.07 & \cellcolor{green!2}37.87 & \cellcolor{red!1}36.27 & \cellcolor{red!1}34.40 & \cellcolor{red!3}33.60 & \cellcolor{green!22}33.60 & \cellcolor{green!28}36.00 & \cellcolor{green!7}35.79 \\
\specialrule{0.10em}{0.05em}{0.05em}
\textsc{Self-Assessment} & \cellcolor{green!6}41.07 & \cellcolor{green!4}37.60 & \cellcolor{green!17}35.20 & \cellcolor{green!4}39.20 & \cellcolor{green!11}43.20 & \cellcolor{green!10}42.40 & \cellcolor{green!8}39.73 & \cellcolor{green!5}37.87 & 20.80 & \cellcolor{green!5}22.93 & \cellcolor{green!7}36.00 \\
\textsc{Native-Tool-Use} & \cellcolor{green!3}39.73 & \cellcolor{green!5}37.87 & \cellcolor{green!7}29.60 & \cellcolor{green!5}40.00 & \cellcolor{green!7}40.80 & \cellcolor{green!7}41.07 & \cellcolor{green!3}37.07 & \cellcolor{red!1}34.40 & \cellcolor{green!6}24.00 & \cellcolor{green!8}24.80 & \cellcolor{green!5}34.93 \\
\textsc{ST} & \cellcolor{green!1}38.13 & \cellcolor{green!1}35.47 & \cellcolor{green!15}33.87 & \cellcolor{green!1}37.87 & \cellcolor{green!5}39.47 & \cellcolor{green!3}38.40 & \cellcolor{green!4}37.60 & \cellcolor{green!1}35.73 & \cellcolor{green!12}27.47 & \cellcolor{green!18}30.67 & \cellcolor{green!6}35.47 \\
\textsc{Boundary-SFT} & \cellcolor{red!9}32.80 & \cellcolor{red!6}32.00 & \cellcolor{green!11}31.60 & \cellcolor{red!5}34.40 & \cellcolor{red!7}32.80 & \cellcolor{red!2}35.60 & 35.20 & \cellcolor{red!8}30.80 & \cellcolor{green!1}21.60 & \cellcolor{green!15}28.80 & \cellcolor{red!1}31.56 \\
\midrule
\textsc{Q-Align} & \cellcolor{red!5}34.93 & \cellcolor{red!3}33.60 & \cellcolor{green!8}29.87 & \cellcolor{red!5}34.40 & \cellcolor{red!2}35.47 & \cellcolor{red!4}34.67 & \cellcolor{red!4}33.07 & \cellcolor{red!3}33.60 & \cellcolor{red!4}18.40 & \cellcolor{red!3}18.40 & \cellcolor{red!2}30.64 \\
\textsc{GRPO} & \cellcolor{green!2}38.93 & \cellcolor{green!3}37.07 & \cellcolor{green!4}27.47 & \cellcolor{green!10}42.67 & \cellcolor{green!10}42.40 & \cellcolor{green!9}41.87 & \cellcolor{green!7}39.20 & \cellcolor{green!8}39.73 & \cellcolor{green!8}25.33 & \cellcolor{green!5}23.20 & \cellcolor{green!7}35.79 \\
\textsc{TAPO} & \cellcolor{green!4}40.00 & \cellcolor{green!3}36.80 & \cellcolor{green!21}37.60 & \cellcolor{green!10}42.67 & \cellcolor{green!7}41.07 & \cellcolor{green!7}41.07 & \cellcolor{green!7}38.93 & \cellcolor{green!7}38.93 & \cellcolor{green!1}21.07 & \cellcolor{green!5}23.20 & \cellcolor{green!7}36.13 \\
\midrule
\rowcolor{gray!12} \textsc{\textbf{Luar}} & \cellcolor{green!5}40.53 & \cellcolor{green!7}38.93 & \cellcolor{green!22}37.87 & \cellcolor{green!6}40.27 & \cellcolor{green!8}41.33 & \cellcolor{green!6}40.27 & \cellcolor{green!9}40.27 & \cellcolor{red!1}34.93 & \cellcolor{green!12}27.47 & \cellcolor{green!30}37.60 & \cellcolor{green!10}\textbf{37.95} \\
\midrule
\textsc{Qwen3-8B} & 36.80 & 36.00 & 29.33 & 34.93 & 34.40 & 36.27 & 33.87 & 33.87 & 21.87 & 23.73 & 32.11 \\
\midrule
\textsc{Full-Translation} & \cellcolor{red!1}36.53 & \cellcolor{red!4}33.60 & \cellcolor{green!13}36.53 & 34.93 & \cellcolor{green!4}36.80 & \cellcolor{green!1}36.80 & \cellcolor{green!2}35.20 & \cellcolor{green!1}34.13 & \cellcolor{green!23}34.93 & \cellcolor{green!21}35.73 & \cellcolor{green!6}35.52 \\
\specialrule{0.10em}{0.05em}{0.05em}
\textsc{Self-Assessment} & \cellcolor{green!8}41.33 & \cellcolor{green!7}40.00 & \cellcolor{green!11}35.73 & \cellcolor{green!6}38.40 & \cellcolor{green!12}41.33 & \cellcolor{green!12}43.20 & \cellcolor{green!8}38.67 & \cellcolor{green!9}38.93 & \cellcolor{red!1}21.60 & \cellcolor{green!7}27.73 & \cellcolor{green!8}36.69 \\
\textsc{Native-Tool-Use} & \cellcolor{red!26}22.13 & \cellcolor{red!26}21.07 & \cellcolor{red!3}27.47 & \cellcolor{red!6}31.47 & \cellcolor{red!10}28.53 & \cellcolor{red!35}16.53 & \cellcolor{red!20}22.67 & \cellcolor{red!8}29.33 & \cellcolor{green!6}25.07 & \cellcolor{green!2}24.80 & \cellcolor{red!13}24.91 \\
\textsc{ST} & \cellcolor{green!7}40.53 & \cellcolor{green!1}36.80 & \cellcolor{green!13}36.53 & \cellcolor{green!5}37.60 & \cellcolor{green!7}38.13 & \cellcolor{green!5}39.20 & \cellcolor{green!4}36.00 & \cellcolor{green!2}35.20 & \cellcolor{green!20}33.33 & \cellcolor{green!19}34.40 & \cellcolor{green!8}36.77 \\
\textsc{Boundary-SFT} & \cellcolor{green!1}37.33 & \cellcolor{red!2}34.67 & \cellcolor{green!6}32.80 & \cellcolor{green!1}35.47 & \cellcolor{green!1}35.20 & \cellcolor{green!1}36.80 & \cellcolor{green!2}35.20 & \cellcolor{red!4}31.73 & \cellcolor{green!6}25.07 & \cellcolor{green!13}30.93 & \cellcolor{green!2}33.52 \\
\midrule
\textsc{Q-Align} & \cellcolor{red!3}34.93 & \cellcolor{red!4}33.60 & \cellcolor{green!7}33.60 & 34.93 & 34.40 & \cellcolor{red!6}33.07 & \cellcolor{red!1}33.60 & \cellcolor{red!6}30.40 & \cellcolor{red!5}18.93 & \cellcolor{red!6}20.53 & \cellcolor{red!2}30.80 \\
\textsc{GRPO} & \cellcolor{green!5}39.47 & \cellcolor{green!5}38.67 & \cellcolor{green!5}32.27 & \cellcolor{green!10}40.80 & \cellcolor{green!11}40.53 & \cellcolor{green!12}43.20 & \cellcolor{green!7}37.87 & \cellcolor{green!7}37.60 & \cellcolor{green!4}24.27 & \cellcolor{green!8}28.53 & \cellcolor{green!7}36.32 \\
\textsc{TAPO} & \cellcolor{green!3}38.40 & \cellcolor{green!1}36.80 & \cellcolor{green!3}31.20 & \cellcolor{green!11}41.07 & \cellcolor{green!5}37.07 & \cellcolor{green!5}38.93 & \cellcolor{green!8}38.67 & \cellcolor{green!1}34.40 & \cellcolor{red!7}17.60 & \cellcolor{red!5}20.80 & \cellcolor{green!2}33.49 \\
\midrule
\rowcolor{gray!12} \textsc{\textbf{Luar}} & \cellcolor{green!10}42.40 & \cellcolor{green!8}40.53 & \cellcolor{green!14}37.60 & \cellcolor{green!16}44.00 & \cellcolor{green!15}43.20 & \cellcolor{green!9}41.60 & \cellcolor{green!9}39.20 & \cellcolor{green!11}40.27 & \cellcolor{green!7}25.87 & \cellcolor{green!15}32.27 & \cellcolor{green!12}\textbf{38.69} \\
\bottomrule
\end{tabular}
\caption{Per-language task accuracy (\%) on PolyMath-Medium. Numbers are mean accuracies over seeds. Bold indicates the best average among methods that do not translate every input. Cell colors indicate changes from the corresponding base-model row, clipped at $\pm$20 points.}
\label{tab:per_language_accuracy_base_relative_colored_polymath_medium}
\end{table*}

\begin{table*}[t]
\centering
\small
\setlength{\tabcolsep}{4.5pt}
\begin{tabular}{lccccccccccc}
\toprule
\textbf{Method} 
& \multicolumn{3}{c}{\textbf{Seen}} 
& \multicolumn{7}{c}{\textbf{Unseen}} 
& \textbf{Avg.} \\
\cmidrule(lr){2-4}\cmidrule(lr){5-11}
& \textbf{ar} & \textbf{th} & \textbf{sw}
& \textbf{en} & \textbf{zh} & \textbf{ko} & \textbf{bn} & \textbf{te} & \textbf{yo} & \textbf{zu}
& \\
\midrule
\textsc{Qwen3-4B} & 24.00 & 24.00 & 15.20 & 23.20 & 22.67 & 25.33 & 24.27 & 17.33 & 9.87 & 11.73 & 19.76 \\
\midrule
\textsc{Full-Translation} & \cellcolor{green!3}25.60 & \cellcolor{green!1}24.80 & \cellcolor{green!18}25.33 & 23.20 & \cellcolor{red!2}21.60 & \cellcolor{red!2}24.00 & 24.27 & \cellcolor{green!8}22.13 & \cellcolor{green!27}25.33 & \cellcolor{green!25}26.13 & \cellcolor{green!8}24.24 \\
\specialrule{0.10em}{0.05em}{0.05em}
\textsc{Self-Assessment} & \cellcolor{green!14}31.73 & \cellcolor{green!4}26.40 & \cellcolor{green!9}20.53 & \cellcolor{green!9}28.53 & \cellcolor{green!9}28.00 & \cellcolor{green!8}29.87 & \cellcolor{green!5}26.93 & \cellcolor{green!16}26.67 & \cellcolor{red!1}9.33 & \cellcolor{red!1}11.47 & \cellcolor{green!7}23.95 \\
\textsc{Native-Tool-Use} & \cellcolor{green!8}28.80 & \cellcolor{green!1}24.80 & \cellcolor{green!4}17.33 & \cellcolor{green!9}28.27 & \cellcolor{green!13}29.87 & \cellcolor{green!8}29.87 & \cellcolor{green!2}25.33 & \cellcolor{green!14}25.07 & \cellcolor{green!6}13.33 & \cellcolor{green!5}14.67 & \cellcolor{green!7}23.73 \\
\textsc{ST} & \cellcolor{green!7}28.00 & \cellcolor{green!3}25.60 & \cellcolor{green!13}22.67 & \cellcolor{green!2}24.53 & \cellcolor{green!8}27.20 & \cellcolor{green!2}26.40 & \cellcolor{green!3}25.87 & \cellcolor{green!8}21.87 & \cellcolor{green!17}19.73 & \cellcolor{green!17}21.33 & \cellcolor{green!8}24.32 \\
\textsc{Boundary-SFT} & \cellcolor{green!1}24.40 & \cellcolor{red!3}22.40 & \cellcolor{green!8}20.00 & \cellcolor{red!1}22.80 & \cellcolor{red!9}17.60 & \cellcolor{red!4}23.20 & \cellcolor{green!2}25.60 & \cellcolor{green!7}21.20 & \cellcolor{green!3}11.60 & \cellcolor{green!11}18.00 & \cellcolor{green!2}20.68 \\
\midrule
\textsc{Q-Align} & \cellcolor{red!2}22.93 & \cellcolor{red!4}21.60 & \cellcolor{green!6}18.40 & \cellcolor{red!4}20.80 & \cellcolor{green!1}23.20 & \cellcolor{red!7}21.33 & \cellcolor{red!3}22.40 & \cellcolor{green!7}21.07 & 9.87 & \cellcolor{red!5}9.07 & \cellcolor{red!1}19.07 \\
\textsc{GRPO} & \cellcolor{green!8}28.53 & \cellcolor{green!1}24.80 & \cellcolor{green!4}17.60 & \cellcolor{green!11}29.33 & \cellcolor{green!6}25.87 & \cellcolor{green!9}30.67 & \cellcolor{green!5}26.93 & \cellcolor{green!7}21.33 & \cellcolor{green!5}12.80 & 11.73 & \cellcolor{green!6}22.96 \\
\textsc{TAPO} & \cellcolor{green!7}28.00 & \cellcolor{green!4}26.40 & \cellcolor{green!11}21.60 & \cellcolor{green!9}28.53 & \cellcolor{green!13}30.13 & \cellcolor{green!2}26.40 & \cellcolor{green!4}26.67 & \cellcolor{green!19}28.00 & \cellcolor{red!1}9.33 & \cellcolor{red!5}9.07 & \cellcolor{green!6}23.41 \\
\midrule
\rowcolor{gray!12} \textsc{\textbf{Luar}} & \cellcolor{green!12}30.67 & \cellcolor{green!9}29.33 & \cellcolor{green!30}32.53 & \cellcolor{green!14}30.93 & \cellcolor{green!11}29.07 & \cellcolor{green!12}32.27 & \cellcolor{green!5}27.20 & \cellcolor{green!14}25.07 & \cellcolor{green!15}18.40 & \cellcolor{green!28}27.73 & \cellcolor{green!15}\textbf{28.32} \\
\midrule
\textsc{Qwen3-8B} & 23.20 & 21.87 & 18.40 & 21.60 & 21.07 & 21.87 & 23.20 & 18.67 & 9.33 & 10.40 & 18.96 \\
\midrule
\textsc{Full-Translation} & \cellcolor{green!1}23.73 & \cellcolor{green!3}23.47 & \cellcolor{green!8}22.93 & 21.60 & \cellcolor{green!6}24.53 & \cellcolor{green!4}24.27 & 23.20 & \cellcolor{green!8}23.20 & \cellcolor{green!28}25.33 & \cellcolor{green!27}25.60 & \cellcolor{green!8}23.79 \\
\specialrule{0.10em}{0.05em}{0.05em}
\textsc{Self-Assessment} & \cellcolor{green!10}29.07 & \cellcolor{green!12}28.80 & \cellcolor{green!10}24.27 & \cellcolor{green!6}25.07 & \cellcolor{green!17}30.93 & \cellcolor{green!14}29.60 & \cellcolor{green!13}30.67 & \cellcolor{green!18}29.07 & \cellcolor{green!6}12.80 & \cellcolor{green!7}14.13 & \cellcolor{green!11}25.44 \\
\textsc{Native-Tool-Use} & \cellcolor{red!5}20.27 & \cellcolor{red!6}18.67 & \cellcolor{green!3}20.27 & 21.60 & \cellcolor{green!5}24.00 & \cellcolor{red!12}14.93 & \cellcolor{red!6}19.73 & \cellcolor{green!8}23.20 & \cellcolor{green!20}20.80 & \cellcolor{green!14}18.67 & \cellcolor{green!2}20.21 \\
\textsc{ST} & \cellcolor{green!7}26.93 & \cellcolor{green!4}24.27 & \cellcolor{green!14}26.13 & \cellcolor{green!5}24.53 & \cellcolor{green!10}26.67 & \cellcolor{green!9}26.93 & \cellcolor{green!1}23.47 & \cellcolor{green!13}25.87 & \cellcolor{green!19}20.00 & \cellcolor{green!24}24.00 & \cellcolor{green!10}24.88 \\
\textsc{Boundary-SFT} & \cellcolor{red!2}22.13 & \cellcolor{green!1}22.40 & \cellcolor{green!5}21.33 & \cellcolor{green!4}24.00 & \cellcolor{green!3}22.67 & 21.87 & \cellcolor{red!1}22.67 & \cellcolor{green!2}19.73 & \cellcolor{green!11}15.47 & \cellcolor{green!10}16.00 & \cellcolor{green!3}20.83 \\
\midrule
\textsc{Q-Align} & \cellcolor{red!5}20.53 & \cellcolor{red!4}19.73 & \cellcolor{green!2}19.47 & \cellcolor{red!4}19.47 & \cellcolor{green!3}22.93 & \cellcolor{red!2}20.80 & \cellcolor{red!2}21.87 & \cellcolor{green!1}19.47 & \cellcolor{red!2}8.00 & \cellcolor{red!2}9.33 & \cellcolor{red!1}18.16 \\
\textsc{GRPO} & \cellcolor{green!16}32.53 & \cellcolor{green!14}29.60 & \cellcolor{green!10}24.27 & \cellcolor{green!19}32.53 & \cellcolor{green!16}30.40 & \cellcolor{green!14}29.87 & \cellcolor{green!9}28.27 & \cellcolor{green!15}27.47 & \cellcolor{green!9}14.67 & \cellcolor{green!7}14.13 & \cellcolor{green!13}26.37 \\
\textsc{TAPO} & \cellcolor{green!8}27.73 & \cellcolor{green!1}22.67 & \cellcolor{green!5}21.33 & \cellcolor{green!11}28.00 & \cellcolor{green!7}25.07 & \cellcolor{green!9}26.93 & \cellcolor{red!1}22.67 & \cellcolor{green!9}23.73 & 9.33 & \cellcolor{red!2}9.33 & \cellcolor{green!5}21.68 \\
\midrule
\rowcolor{gray!12} \textsc{\textbf{Luar}} & \cellcolor{green!10}28.80 & \cellcolor{green!10}27.73 & \cellcolor{green!19}29.33 & \cellcolor{green!16}30.67 & \cellcolor{green!14}29.33 & \cellcolor{green!17}31.73 & \cellcolor{green!10}29.07 & \cellcolor{green!15}27.20 & \cellcolor{green!20}20.80 & \cellcolor{green!24}24.00 & \cellcolor{green!16}\textbf{27.87} \\
\bottomrule
\end{tabular}
\caption{Per-language task accuracy (\%) on PolyMath-High. Numbers are mean accuracies over seeds. Bold indicates the best average among methods that do not translate every input. Cell colors indicate changes from the corresponding base-model row, clipped at $\pm$20 points.}
\label{tab:per_language_accuracy_base_relative_colored_polymath_high}
\end{table*}

\begin{table*}[t]
\centering
\small
\setlength{\tabcolsep}{4.5pt}
\begin{tabular}{lccccccccccc}
\toprule
\textbf{Method} 
& \multicolumn{3}{c}{\textbf{Seen}} 
& \multicolumn{7}{c}{\textbf{Unseen}} 
& \textbf{Avg.} \\
\cmidrule(lr){2-4}\cmidrule(lr){5-11}
& \textbf{ar} & \textbf{th} & \textbf{sw}
& \textbf{en} & \textbf{zh} & \textbf{ko} & \textbf{bn} & \textbf{te} & \textbf{yo} & \textbf{zu}
& \\
\midrule
\textsc{Qwen3-4B} & 72.11 & 71.73 & 50.71 & 75.10 & 71.34 & 72.37 & 70.95 & 67.32 & 42.28 & 40.99 & 63.49 \\
\midrule
\textsc{Full-Translation} & \cellcolor{green!5}74.71 & \cellcolor{green!6}75.10 & \cellcolor{green!35}72.11 & 75.10 & \cellcolor{green!8}76.13 & \cellcolor{green!4}74.58 & \cellcolor{green!9}76.01 & \cellcolor{green!10}73.02 & \cellcolor{green!35}64.33 & \cellcolor{green!35}68.35 & \cellcolor{green!17}72.94 \\
\specialrule{0.10em}{0.05em}{0.05em}
\textsc{Self-Assessment} & \cellcolor{green!2}73.15 & \cellcolor{green!1}72.24 & \cellcolor{red!1}50.06 & \cellcolor{red!1}74.84 & \cellcolor{green!6}74.71 & \cellcolor{green!1}73.02 & \cellcolor{green!1}71.73 & \cellcolor{green!2}68.35 & \cellcolor{red!1}41.63 & \cellcolor{green!5}43.84 & \cellcolor{green!2}64.36 \\
\textsc{Native-Tool-Use} & \cellcolor{green!2}73.28 & \cellcolor{green!4}73.93 & \cellcolor{green!7}54.60 & \cellcolor{green!2}76.52 & \cellcolor{red!3}69.65 & \cellcolor{green!1}72.63 & \cellcolor{green!2}72.24 & \cellcolor{green!1}67.96 & \cellcolor{green!7}46.43 & \cellcolor{green!17}50.45 & \cellcolor{green!4}65.77 \\
\textsc{ST} & \cellcolor{green!3}73.93 & \cellcolor{green!5}74.84 & \cellcolor{green!35}71.47 & \cellcolor{green!1}75.75 & \cellcolor{green!7}75.36 & \cellcolor{green!2}73.54 & \cellcolor{green!2}72.37 & \cellcolor{green!8}71.73 & \cellcolor{green!34}61.74 & \cellcolor{green!35}64.98 & \cellcolor{green!14}\textbf{71.57} \\
\textsc{Boundary-SFT} & \cellcolor{red!1}71.60 & \cellcolor{red!2}70.43 & \cellcolor{green!6}53.89 & \cellcolor{red!6}71.60 & \cellcolor{red!10}65.37 & \cellcolor{red!1}71.79 & \cellcolor{red!2}69.84 & \cellcolor{red!8}62.65 & \cellcolor{green!1}42.41 & \cellcolor{green!27}56.42 & \cellcolor{green!1}63.60 \\
\midrule
\textsc{Q-Align} & \cellcolor{red!7}68.09 & \cellcolor{red!3}69.78 & \cellcolor{green!3}52.27 & \cellcolor{red!5}72.37 & \cellcolor{red!2}70.04 & \cellcolor{red!8}67.57 & \cellcolor{red!7}66.67 & \cellcolor{red!6}63.81 & \cellcolor{red!19}31.26 & \cellcolor{red!9}36.06 & \cellcolor{red!6}59.79 \\
\textsc{GRPO} & \cellcolor{red!6}68.48 & \cellcolor{red!5}68.74 & \cellcolor{red!7}46.95 & \cellcolor{red!3}73.41 & \cellcolor{green!2}72.37 & \cellcolor{red!3}70.56 & \cellcolor{red!3}69.13 & \cellcolor{red!3}65.63 & \cellcolor{red!6}38.78 & \cellcolor{red!2}40.08 & \cellcolor{red!4}61.41 \\
\textsc{TAPO} & \cellcolor{red!6}68.87 & \cellcolor{red!1}71.34 & \cellcolor{green!8}55.25 & \cellcolor{red!1}74.58 & \cellcolor{green!1}71.98 & \cellcolor{red!5}69.39 & \cellcolor{red!6}67.32 & \cellcolor{red!4}64.85 & \cellcolor{red!17}32.81 & \cellcolor{red!9}36.06 & \cellcolor{red!4}61.25 \\
\midrule
\rowcolor{gray!12} \textsc{\textbf{Luar}} & \cellcolor{red!1}71.47 & \cellcolor{green!4}74.19 & \cellcolor{green!32}69.26 & \cellcolor{green!1}75.36 & \cellcolor{green!1}72.11 & 72.37 & \cellcolor{red!1}70.30 & 67.32 & \cellcolor{green!8}46.69 & \cellcolor{green!35}65.24 & \cellcolor{green!9}68.43 \\
\midrule
\textsc{Qwen3-8B} & 76.52 & 76.91 & 56.81 & 76.01 & 75.62 & 74.58 & 73.93 & 69.00 & 42.80 & 44.62 & 66.68 \\
\midrule
\textsc{Full-Translation} & \cellcolor{green!5}79.38 & \cellcolor{green!5}80.03 & \cellcolor{green!33}75.75 & 76.01 & \cellcolor{green!6}78.86 & \cellcolor{green!5}77.69 & \cellcolor{green!5}76.78 & \cellcolor{green!14}76.91 & \cellcolor{green!35}67.32 & \cellcolor{green!35}70.17 & \cellcolor{green!16}75.89 \\
\specialrule{0.10em}{0.05em}{0.05em}
\textsc{Self-Assessment} & 76.52 & \cellcolor{green!1}77.56 & \cellcolor{green!8}61.48 & \cellcolor{green!4}78.47 & \cellcolor{green!4}77.95 & \cellcolor{green!2}75.75 & \cellcolor{green!2}74.84 & \cellcolor{green!6}72.63 & \cellcolor{green!7}46.69 & \cellcolor{green!6}47.86 & \cellcolor{green!4}68.98 \\
\textsc{Native-Tool-Use} & \cellcolor{red!3}74.58 & \cellcolor{red!21}64.98 & \cellcolor{green!19}67.57 & \cellcolor{red!2}74.97 & \cellcolor{red!5}72.50 & \cellcolor{red!2}73.41 & \cellcolor{green!5}77.04 & \cellcolor{green!3}70.82 & \cellcolor{green!35}63.81 & \cellcolor{green!35}66.80 & \cellcolor{green!7}70.65 \\
\textsc{ST} & \cellcolor{green!4}78.86 & \cellcolor{green!3}78.86 & \cellcolor{green!30}74.06 & \cellcolor{green!7}80.16 & \cellcolor{green!5}78.73 & \cellcolor{green!5}77.56 & \cellcolor{green!7}77.95 & \cellcolor{green!10}74.45 & \cellcolor{green!35}65.63 & \cellcolor{green!35}67.32 & \cellcolor{green!15}\textbf{75.36} \\
\textsc{Boundary-SFT} & \cellcolor{red!5}73.80 & \cellcolor{red!8}72.50 & \cellcolor{red!2}55.38 & \cellcolor{red!3}74.45 & \cellcolor{red!8}70.82 & \cellcolor{red!3}72.89 & \cellcolor{red!6}70.69 & \cellcolor{red!1}68.87 & \cellcolor{green!6}46.43 & \cellcolor{green!12}51.75 & \cellcolor{red!2}65.76 \\
\midrule
\textsc{Q-Align} & \cellcolor{red!7}72.24 & \cellcolor{red!5}73.93 & \cellcolor{red!1}56.03 & \cellcolor{green!1}76.78 & \cellcolor{red!4}73.15 & \cellcolor{red!4}72.11 & \cellcolor{red!7}69.91 & \cellcolor{red!1}68.48 & \cellcolor{red!11}36.45 & \cellcolor{red!6}41.37 & \cellcolor{red!5}64.05 \\
\textsc{GRPO} & \cellcolor{red!1}76.26 & \cellcolor{red!1}76.39 & \cellcolor{red!2}55.38 & \cellcolor{green!5}78.99 & \cellcolor{green!1}75.88 & \cellcolor{green!1}75.23 & \cellcolor{green!1}74.58 & \cellcolor{green!6}72.24 & \cellcolor{green!1}43.19 & 44.62 & \cellcolor{green!1}67.28 \\
\textsc{TAPO} & \cellcolor{red!14}68.48 & \cellcolor{red!32}58.75 & \cellcolor{red!12}50.06 & \cellcolor{red!4}73.67 & \cellcolor{red!22}62.91 & \cellcolor{red!18}64.33 & \cellcolor{red!18}63.68 & \cellcolor{red!11}62.52 & \cellcolor{red!25}28.40 & \cellcolor{red!19}33.98 & \cellcolor{red!17}56.68 \\
\midrule
\rowcolor{gray!12} \textsc{\textbf{Luar}} & \cellcolor{red!1}75.75 & \cellcolor{red!1}76.39 & \cellcolor{green!15}65.24 & \cellcolor{red!1}75.36 & \cellcolor{green!1}75.88 & \cellcolor{green!1}75.23 & \cellcolor{red!4}71.47 & \cellcolor{green!2}70.30 & \cellcolor{green!20}54.09 & \cellcolor{green!29}61.09 & \cellcolor{green!6}70.08 \\
\bottomrule
\end{tabular}
\caption{Per-language task accuracy (\%) on MMLU-ProX-Lite. Numbers are mean accuracies over seeds. Bold indicates the best average among methods that do not translate every input. Cell colors indicate changes from the corresponding base-model row, clipped at $\pm$20 points.}
\label{tab:per_language_accuracy_base_relative_colored_mmlu_prox_lite}
\end{table*}

%% file: latex/table/appendix_within_language_call_rates.tex
\begin{table*}[t]
\centering
\small
\caption{Within-language translator-call rates of \textsc{Luar} with Qwen3-4B, pooled across evaluation datasets. Direct-solvable and translation-rescued groups are defined by sampled base-model outcomes. A dash indicates that the group is empty.}
\label{tab:within_language_call_rates}
\setlength{\tabcolsep}{8pt}
\begin{tabular}{lrrrrr}
\toprule
& \multicolumn{2}{c}{\textbf{Translator-call rate (\%)}} & & \multicolumn{2}{c}{\textbf{Number of queries}} \\
\cmidrule(lr){2-3}\cmidrule(lr){5-6}
\textbf{Language} & \textbf{Direct-solvable} & \textbf{Translation-rescued} & \textbf{Gap (pts)} & \textbf{Direct-solvable} & \textbf{Translation-rescued} \\
\midrule
\rowcolor{gray!12}Overall & 8.3 & 72.1 & +63.7 & 2,488 & 476 \\
\midrule
Arabic (ar) & 0.3 & 0.0 & $-0.3$ & 312 & 4 \\
Thai (th) & 0.0 & 0.0 & 0.0 & 307 & 12 \\
Swahili (sw) & 80.3 & 98.0 & +17.8 & 152 & 101 \\
English (en) & 0.0 & --- & --- & 342 & 0 \\
Chinese (zh) & 0.0 & 0.0 & 0.0 & 316 & 7 \\
Korean (ko) & 0.0 & 0.0 & 0.0 & 324 & 4 \\
Bengali (bn) & 2.3 & 0.0 & $-2.3$ & 301 & 13 \\
Telugu (te) & 3.8 & 7.7 & +3.9 & 264 & 26 \\
Yoruba (yo) & 22.8 & 61.3 & +38.5 & 92 & 150 \\
Zulu (zu) & 59.0 & 94.3 & +35.4 & 78 & 159 \\
\bottomrule
\end{tabular}
\end{table*}